\documentclass[a4paper,twoside]{article}

\usepackage{epsfig}
\usepackage{subcaption}
\usepackage{calc}
\usepackage{amssymb}
\usepackage{amstext}
\usepackage{amsmath}
\usepackage{amsthm}
\usepackage{multicol}
\usepackage{pslatex}
\usepackage{apalike}
\usepackage{todonotes}
\usepackage{hyperref}
\usepackage{SCITEPRESS}     % Please add other packages that you may need BEFORE the SCITEPRESS.sty package.

\usepackage{array}
\newcolumntype{L}[1]{>{\raggedright\let\newline\\\arraybackslash\hspace{0pt}}m{#1}}
\newcolumntype{C}[1]{>{\centering\let\newline\\\arraybackslash\hspace{0pt}}m{#1}}
\newcolumntype{R}[1]{>{\raggedleft\let\newline\\\arraybackslash\hspace{0pt}}m{#1}}

\begin{document}
\title{Reconstruction of Convex Polytope Compositions from 3D Point-clouds}

\author{\authorname{Markus Friedrich\sup{1}, Pierre-Alain Fayolle\sup{2}}
\affiliation{\sup{1}Institute for Computer Science, LMU Munich, Oettinge9stra{\ss}e 67, Munich, Germany}
\affiliation{\sup{2}The University of Aizu, Ikki machi, Aizu-Wakamatsu, Japan}
\email{markus.friedrich@ifi.lmu.de, fayolle@u-aizu.ac.jp}
}

%\author{\authorname{Anonymous\sup{1}}
%\affiliation{\sup{1}Anonymous Institute}
%\email{anonymous@anonymous.com}
%}

\abstract{
Reconstructing a composition (union) of convex polytopes that perfectly fits the corresponding input point-cloud is a hard optimization problem with interesting applications in reverse engineering and rigid body dynamics simulations.
We propose a pipeline that first extracts a set of planes, then partitions the input point-cloud into weakly convex clusters and finally generates a set of convex polytopes as the intersection of fitted planes for each partition.
Finding the best-fitting convex polytopes is formulated as a combinatorial optimization problem over the set of fitted planes and is solved using an Evolutionary Algorithm. 
For convex clustering, we employ two different methods and detail their strengths and weaknesses in a thorough evaluation based on multiple input data-sets. 
}

\onecolumn \maketitle \normalsize \setcounter{footnote}{0} \vfill

\section{\uppercase{Introduction}}\label{sec:introduction}
This work deals with the problem of reconstructing a solid object from an input 3D point-cloud, where the solid object is represented as a collection (union) of convex polytopes (represented as an intersection of planar half-spaces). 
We are dealing with the particular case where the input 3D point-cloud describes a single, or a few, at most, objects. Larger scenes, made of multiple objects, can be first decomposed by classification or semantic segmentation. 

Potential applications can be found in reverse engineering, or reconstruction, of buildings from 3D point-clouds, see for example \cite{Musialski_cgf13} or in the field of numerical physics simulations, such as in the simulation of rigid body dynamics, see for example \cite{coumans2019}, where the convex decomposition of a solid can increase efficiency, since the collision of convex bodies can be efficiently determined \cite{Gilbert_jra88}.  

The problem of reconstructing a set of convex polytopes from a 3D point-cloud is difficult to solve since it heavily relies on the robust detection and exact fitting of planes in the unstructured input point-cloud, and on finding a correct mapping of these planes in a set of convex polytopes.
The latter is a difficult combinatorial problem since neither the number of resulting convex polytopes nor the sets of planes that form particular polytopes are known in advance. 
Our approach relies on a pre-segmentation step performed by a deep neural network, followed by fitting planes using a RANSAC-based model fitter. 
Then, a clustering of the input point-cloud into weakly convex parts is conducted and the clusters are used by an Evolutionary Algorithm to form a collection of convex polytopes. 
Our main contributions are: 
\begin{itemize}
    \item A detailed description and evaluation of a full pipeline for the detection and fitting of convex polytopes in an unstructured 3D point-cloud. 
    \item A detailed comparison of clustering methods to group points in (weakly) convex clusters. 
    \item An efficient Evolutionary Algorithm (EA) to form a collection of convex polytopes from clusters of points. 
\end{itemize}

The rest of this paper is organized as follows: 
First we discuss works related to our approach in Section~\ref{sec:related}, followed by an introduction of the basic concepts used in the rest of the paper in Section~\ref{sec:background}. 
In Section~\ref{sec:concept}, we describe our reconstruction pipeline in detail. 
It is followed by its evaluation in Section~\ref{sec:evaluation}. 
Finally, the paper ends with a brief conclusion and a discussion of potential future directions of work (Section~\ref{sec:conclusion}).

\section{\uppercase{Related Works}}
\label{sec:related}
The problems of segmentation, primitive detection and fitting are well studied in computer graphics, computer vision, computer aided design and related engineering domains, see for example this survey on primitive detection \cite{Kaiser:2019:GeoPrimFitSurvey} and the references therein. 
We deal in this work with the problem of reconstructing a solid from a 3D point-cloud as a composition (union) of convex polytopes. 
Thus, we are interested in related works considering problems such as: plane detection and fitting, cuboid or general polytope detection and fitting, among others. In the following, we list the works most relevant to these problems. 

Segmentation, primitive detection and fitting are necessary steps in the field of reverse engineering 3D data, which is the process of recovering a computer model of a 3D shape from acquired data. See, for example, \cite{VBK98,BV04} and the references therein. 
Approaches in reverse engineering are, however, not limited to the fitting of planar patches, but deal also with higher order patches common in industrial design. On the other hand, fitted planar patches do not have to be arranged into collections of cuboids or convex polytopes in these approaches unlike the problem that we are dealing with. 

A popular technique for fitting models to data (including noisy data) is RANSAC \cite{fischler1981random}, as well as its numerous variants. 
The efficient RANSAC method introduced in \cite{schnabel2007efficient} is a fast RANSAC-based approach for detecting and fitting primitives of different types (plane, cylinder, sphere, cone) in a 3D point-cloud. 
The approach was further improved in \cite{LWCSCOM11} by enforcing additional constraints during the fitting process, such as the fact that two planes are parallel or perpendicular. The addition of these constraints allow for a more robust fitting of the primitives at the cost of a less efficient approach. 
While the efficient RANSAC approach \cite{schnabel2007efficient} deals with unbounded primitives (plane, infinite cylinder), the method described in \cite{friedrich2020hybrid} uses additional steps to generate solid primitives. 
Our approach also uses RANSAC as one of its steps. However, we apply RANSAC to a pre-clustered point cloud, which allows us to make the process more robust and less parameter sensitive. 
In addition, unlike \cite{schnabel2007efficient} that fits infinite planes, we generate convex polytopes by combining the initially fitted planes. Note that our approach deals with general convex polytopes unlike \cite{friedrich2020hybrid} that is limited to cuboids.  

The efficient detection and fitting of planes in 3D point-clouds is a necessary step for the reconstruction of buildings. See for example \cite{monszpart2015rapter,oesau_cgf16} and the references therein. 
These planes can then be combined to form cuboids \cite{xiao2014,li2016boxfitting} or more complex polyhedral shapes \cite{nan2017polyfit}. 
In the work \cite{xiao2014}, the authors propose a method to reconstruct museums by fitting cuboids to the input data and by combining them using a CSG (Constructive Solid Geometry) expression. 
The method described in \cite{li2016boxfitting} assumes that all fitted planes are perpendicular to one of the three dominant directions. This allows to recast the combinatorial problem of combining planes to form cuboids into an energy minimization problem that can be solved using graph-cut optimization. Unlike these works, we deal with the problem of forming a minimal (or at least as small as possible) set of general convex polytopes describing the solid, and are not limited to cuboids. 
The method presented in \cite{nan2017polyfit} relaxes the constraint that planes need to be perpendicular to the three main directions of the data, and instead deals with the minimization of a binary linear problem that is solved with an off-the-shelf solver. This approach forms one polytope (not necessarily convex) for a given input point-cloud. On the other hand, we deal with the problem of finding a set of convex polytopes describing the input point-cloud. 

In recent years, techniques from machine learning, such as deep neural networks, have become popular tools for problems of classification, segmentation or fitting/discovery of models from 3D point-clouds. 
\\
The approach described in \cite{tulsiani2017learning} uses a deep neural network to approximate an input 3D shape by predicting a collection of cuboids. 
\\
A method for learning a convex shape decomposition from an input image, called CvxNet, is introduced in \cite{deng2019cvxnets}. The method consists of training a deep neural network that defines a solid object as a union of convex shapes, where each convex shape is represented by a combination of planar half-spaces. The input is assumed to be an image (RGB or depth image), while we work with unstructured point-clouds. Furthermore, they assume a fixed number of polytopes (i.e. their network always output the same number of convex polytopes). 
\\
A related approach for learning a BSP tree (Binary Space Partition) from an input image or an input voxel is proposed in \cite{chen2019bsp}. Similar to our approach it outputs a collection of convex polytopes describing an object. However, the approach is based on learning from a collection of shapes belonging to a set of categories and is thus restricted to process objects belonging to these same categories. 

\section{\uppercase{Background}}
\label{sec:background}

\subsection{Evolutionary Algorithms}
Evolutionary Algorithms are population-based, iterative meta heuristics for solving mainly combinatorial optimization problems.
The optimization process starts with the creation of a population of randomly generated solution candidates. 
All candidates in the population are then ranked based on a objective (or fitness) function which is the formal description of the objective that should be optimized for.
Based on the ranking, a subset of high-ranked solution candidates are selected to form the next iteration's population. 
The rest of the population is filled with stochastic variations of selected individuals from the old population. 
These variations are described in form of so-called mutation and crossover operators and are highly domain-specific.
Whereas mutation operators usually alter a single individual randomly, crossover operators exchange random parts between two or more individuals.
Variation operators are applied with a certain probability (also called mutation rate and crossover rate).
The execution ends if a certain termination criteria is met (e.g. maximum number of iterations or a target quality has been reached).
The main advantage of Evolutionary Algorithms is their flexibility: 
Solution candidate representation, objective function as well as variation operators can be tailored to specific application domains.  
Furthermore, the objective function does not have to be differentiable like in gradient-based optimization algorithms. 

\subsection{Convex Polytopes}
A 3D convex polytope (in the following convex polytope or polytope w.l.o.g.) is a special-case of a polyhedron with the additional property that its surface encloses a convex subset of the Euclidean space.
A convex polytope can either be described by the intersection of a set of planar half-spaces (H-representation) or by its extreme points (V-representation) which are essentially the vertices of its hull. 
Our approach forms convex polytopes out of planes (H-representation) but also needs the V-representation for volume discretization (see Section~\ref{ch:ga}).
The transformation between H- and V- representation can be done using the Double Description method \cite{Fukuda1995DoubleDM}.
In addition, the signed distance from a 3D point $x$ to the surface of the convex polytope is needed  (see Equation \ref{eq:fgeo}) which is
\begin{equation}
 d(x) =  \min(\{ \text{dot}(p_n, p_o - x): (p_o, p_n) \in P \}),
\end{equation}
where $P$ is the set of planes forming the convex polytope, $p_n$ is a plane's normalized normal, $p_o$ an arbitrary point on the plane and $\text{dot}(\cdot,\cdot)$ is the scalar product of two 3D vectors. 

\section{\uppercase{Pipeline}}
\label{sec:concept}
\begin{figure}[!h!tbp]
	\centering
	\includegraphics[width=0.9\linewidth]{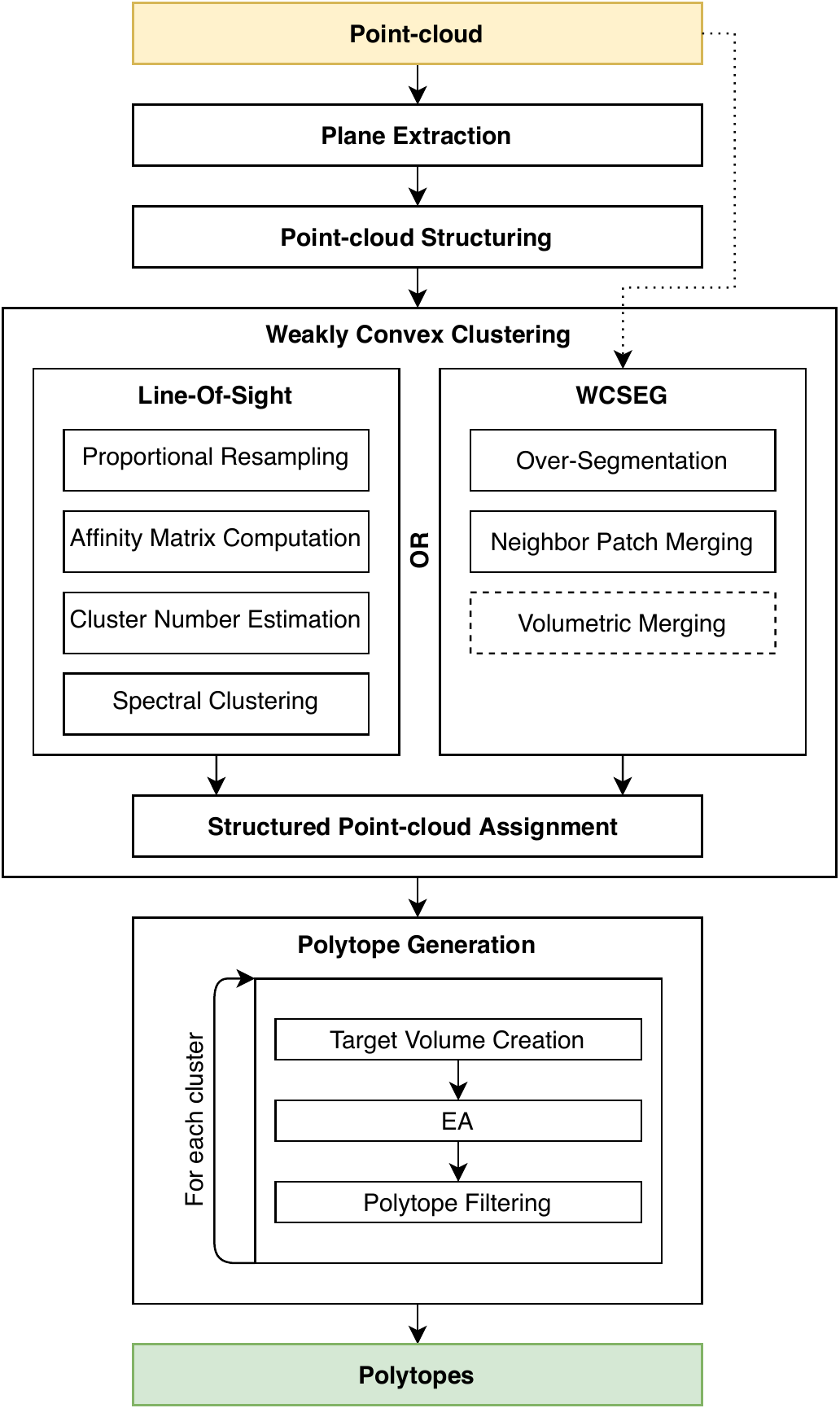}
	\caption{The proposed polytope reconstruction pipeline.}
	\label{fig:pipeline}
\end{figure}

The polytope reconstruction pipeline consists of multiple steps as depicted in Fig.~\ref{fig:pipeline}. 
It starts with a 3D point-cloud as input and ends with the resulting set of convex polytopes as output. 
First, planes are fitted to the input point-cloud (Section~\ref{ch:plane_extraction}).
Then the point-cloud is structured in order to produce a plane-neighborhood graph (Section~\ref{ch:structuring}). 
The point-cloud is then clustered in weakly convex parts (Section~\ref{ch:clustering}). 
Finally, an Evolutionary Algorithm is run on each cluster and its corresponding planes to form a set of convex polytopes (Section~\ref{ch:ga}). 
%The intermediate steps are depicted in Fig.~\ref{fig:pipeline} and described in detail in the following sections.

\subsection{Plane Extraction}
\label{ch:plane_extraction}
In the first step, planes are fitted to the input point-cloud.
We use a clustered variant of the efficient RANSAC approach \cite{schnabel2007efficient} as described in \cite{friedrich2020hybrid}.
It starts with a per-point prediction of primitive types using a deep neural network which was trained with focus on noise and outlier robustness.
This is followed by a DBSCAN clustering \cite{Ester1996DBSCAN} based on the point coordinate, normal and primitive type. 
Finally, parameters are extracted using RANSAC for each cluster and the resulting primitives are merged. 
The additional clustering provably increases fitting robustness \cite{friedrich2020hybrid}. 
This step results in a set of planes $P$ and a mapping $f$ that associates each plane with a subset of surface points from $O$, $f_p: P\to \mathcal{P}(O)$, where $\mathcal{P}$ is the power set operator. 
The result of this step on some test models is illustrated in Fig.~\ref{fig:fit_planes}, where points are colored based on the fitted plane they belong to. 

\subsection{Point-cloud Structuring} \label{ch:structuring}
We apply a point structuring mechanism to the input point-cloud $O$ as proposed in \cite{lafarge_eg13}.  
First, points are projected within a given $\epsilon$ on an occupancy grid located on the surface of their corresponding plane (the grid cell size is $\sqrt{2}\epsilon$).
The occupied cell centers are added to the result point set $O_s$ and marked as points of type 'planar'.
Then, the plane neighborhood graph $G_N = (P, N)$ is extracted, with planes $P$ as vertices and edges $N$ whenever two planes are neighbors (or adjacent). 
Two planes are considered to be adjacent if at least two points of each plane share an edge in the $k$ nearest neighbor ($k$-NN) graph of the input point-cloud $O$.
Based on $G_N$, creases and corner points are extracted, and creases are uniformly sampled using a sampling distance of $2\epsilon$.
Finally, corner and sampled crease points are added to $O_s$ (see Fig.~\ref{fig:structured} for labeled result point sets).
Please note, that we are only interested here in the plane neighborhood graph $G_N$ and the structured point set $O_s$, but not in the point labeling ('planar', 'crease' or 'corner').
$O_s$ is furthermore free of noise and outliers.  

\subsection{Weakly Convex Clustering}\label{ch:clustering}
For point-cloud decomposition in convex or almost (weakly) convex clusters, we have experimented with two methods with different performances. 
Both approaches result in a set of clusters $C$ with each cluster $c \in C$ containing an associated point-cloud $O_c$ and a set of associated planes $P_c$.
\\
The results obtained from these weak clustering approaches on some of our test models are shown in Fig.~\ref{fig:clustering_result}. 

\subsubsection{Line-of-Sight (LoS)}\label{ch:los_clustering}
\begin{figure}[!h!tbp]
	\centering
	\includegraphics[width=0.8\linewidth]{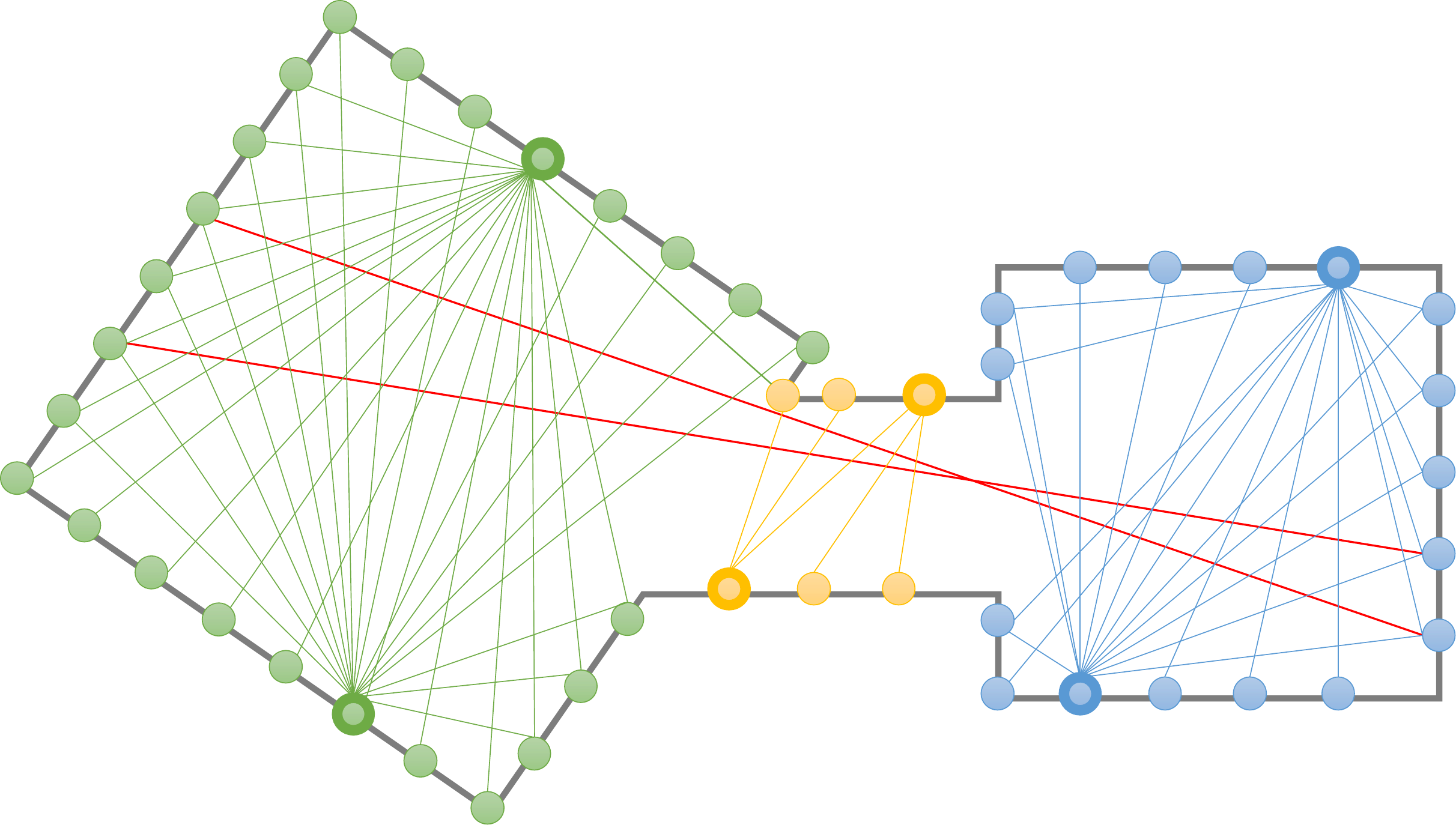}
	\caption{A set of sample points $O$ (convex clusters in green, orange and blue) with exemplary line-of-sights $LoS(O)$ for each convex cluster (framed circles) of the input model. Red lines show inter-partition line-of-sights and grey lines correspond to a piece-wise linear approximation of the surface inferred from the samples. Within a convex cluster, the line-of-sights form a fully connected graph with the cluster samples as nodes. 
	Note that when two mutually visible points belong to the same plane, we do not draw the corresponding line-of-sight for illustration purpose.}
	\label{fig:convex_sample}
\end{figure}

The line-of-sight approach for point-cloud clustering is a variant of the method proposed in \cite{asafi2013weak}. 
The main idea is to extract a graph $G_V = (O, LoS(O))$ with its vertices being the points of the input point-cloud $O$ and its edges expressed by the mapping $LoS: O\to O \times O$ with a point set as domain whose image contains the set of mutually visible point pairs (see Fig.~\ref{fig:convex_sample} as an example). 
This so-called visibility graph has fully or almost fully connected components (cliques) where the corresponding model partition is convex or weakly convex. 
The problem of finding all maximal cliques in a graph is NP hard, but we can use Spectral Clustering as an approximation.    
Thus, in order to extract these convex model parts, the graph is clustered, resulting in a set of point-clouds - one for each convex part.
The different steps necessary for the sketched process of weakly convex clustering are further detailed below. 
\\
\textbf{Proportional Resampling.}
Since the line-of-sight computations have quadratic complexity with respect to the input point-cloud size, the input point-cloud $O_s$ is re-sampled using Farthest Point Sampling (FPS).  %\cite{GONZALEZ1985}.
In order to maintain relative point density, as established by the point structuring step (Section~\ref{ch:structuring}), for large surface areas, FPS is applied for points of each plane separately.
The number of remaining points per plane $k_i$ is 
\begin{equation}
k_i = k\frac{|f_{sr}(p_i)|}{|O_s|}, \; i \in \{1, ..., |P|\},
\end{equation}
where $k$ is the user-controlled accumulated size of all per-plane output point-clouds (we used $k=3000$ in our experiments) and $f_{sr}$ is the mapping between planes in $P$ and points in $O_s$.
This step results in a thinned-out point-cloud $O_{sr}$.
\\
\textbf{Affinity Matrix Computation.}
Spectral Clustering is performed on the so-called affinity matrix $A$ which is the adjacency matrix of the visibility graph $G_v$ and reads: 
\begin{equation}
    A_{i,j} = 
    \begin{cases}
1, & \text{if } (v_i, v_j) \in LoS(O_{sr}) \\
0, & \text{otherwise} 
\end{cases},
\end{equation}
where the line-of-sights $LoS(O_{sr})$ necessary for the affinity matrix are computed on the structured and thinned-out input point-cloud $O_{sr}$. 
A line-of-sight between two points exists if the segment that connects these two points does not intersect with the model's surface - thus, both points are visible from each other. 
Since we don't have a surface but only a set of points $O_{sr}$ and plane primitives $P$, it is necessary to approximate the surface in order to perform the necessary intersection tests. 
This is done by projecting the set of points associated to a given plane on that plane and computing the $2$D Alpha Shape \cite{Edelsbrunner1994} of these points. 
This results in a piece-wise triangulated surface reconstruction for each plane. 
%(see Fig.~\ref{fig:alpha_shape} for an example). 
The visibility check for a point-point segment iterates through all planes and if the segment intersects with a plane, it performs an additional intersection test with each triangle associated to that plane. 
If an intersection is detected, there is no line-of-sight between the two points.
%\begin{figure}[!h!tbp]
%	\centering
%	\includegraphics[width=0.7\linewidth]{figures/alpha_shape.png}
%	\caption{Mesh consisting of all per-plane Alpha Shapes for model M$1$.}
%	\label{fig:alpha_shape}
%\end{figure}
\\
In \cite{asafi2013weak} a more efficient technique for the computation of $A$ is proposed:
From each point $o$ in $O_s$, rays (around $50$-$100$) in the opposite direction of the point's normal and with a certain random direction deviation (maximum of $30$ degrees) are intersected with the surface approximation. 
The point from $O_s$ which is closest to the first ray-surface intersection is considered to be visible from $o$. 
However, our experiments revealed that this method lead to affinity matrices which are too sparse and thus to a low-quality clustering for our data-sets.
\\
\textbf{Spectral Clustering.}
Given the affinity matrix $A$, the degree matrix $D$ is the diagonal matrix with diagonal element $d_i=\sum_j A_{i,j}$.
\\
The un-normalized Laplacian matrix is given by $L=D-A$ (this corresponds to the graph Laplacian matrix when $A$ is the graph adjacency matrix).
\\
There are two commonly used expressions for the normalized Laplacian matrices: 
\begin{align}
    L_{sym} &= D^{-1/2} L D^{1/2} \\
    L_{rw} &= D^{-1} L
\end{align}
$L_{sym}$ is the symmetric normalized Laplacian and $L_{rw}$ is the so-called random walk normalized Laplacian. 
\\
Spectral Clustering is performed by an eigen-analysis of the normalized Laplacian, followed by a $k$-Means clustering of the eigenvectors.
\\
In our experiments, we have found no particular differences between using $L_{sym}$ or $L_{rw}$ in the eigen-analysis. 
Additional details and references on Spectral Clustering are provided in \cite{Luxburg07}. 
\\
\textbf{Estimation of the number of clusters.}
Our implementation of Spectral Clustering uses $k$-Means for clustering the first $k$ eigenvectors of the graph Laplacian corresponding to $A$.
The number of clusters $k$ is usually unknown and highly data specific. 
We tried several techniques to estimate $k$ (such as a density-based clustering technique like DBSCAN \cite{Ester1996DBSCAN} instead of $k$-Means, or finding gaps in the sequence of eigenvalues of the graph Laplacian), but none worked consistently in our tests. 
Instead, we obtained the best results with an idea introduced in \cite{asafi2013weak}:
Spectral Clustering is performed for different values of $k$. 
For each clustering using $k$ clusters ($|C_k|=k$) the clustering quality is measured by 
\begin{equation}
Q(C_k) = \frac{1}{|O_c|^2}\sum_{c\in C_k} |LoS(O_c)| + \alpha |\overline{LoS}(O_c, \overline{O_c})|,
\end{equation}
where the image of the mapping $\overline{LoS}: O_1,O_2 \to O_1 \times O_2$ contains pairs of points $(o_1,o_2), o_1 \in O_1, o_2 \in O_2$ that are not mutually visible. 
$\overline{O_c}$ is used here as an abbreviation for the point set $O_{sr} \setminus O_c$.
The user-defined weighting parameter $\alpha$ was set to $1$ in our experiments. 
The clustering $C_k$ with the highest quality measure $Q(C_k)$ is selected.

\subsubsection{Weakly Convex Segmentation (WCSEG)}
\begin{figure}[!h!t!bp]
	\centering
	\includegraphics[width=0.45\linewidth]{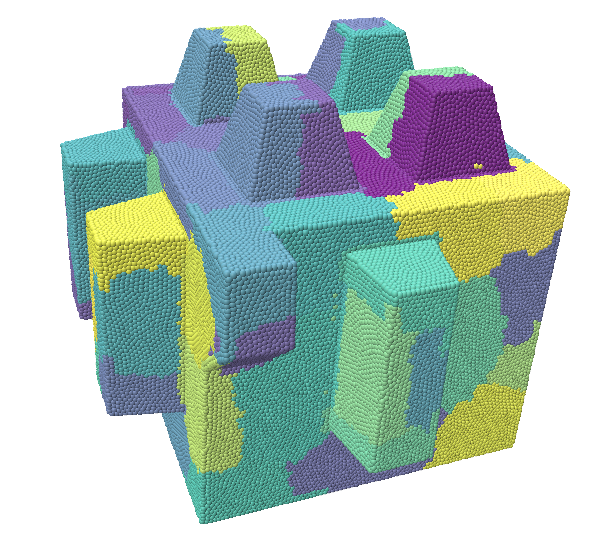}
	\includegraphics[width=0.45\linewidth]{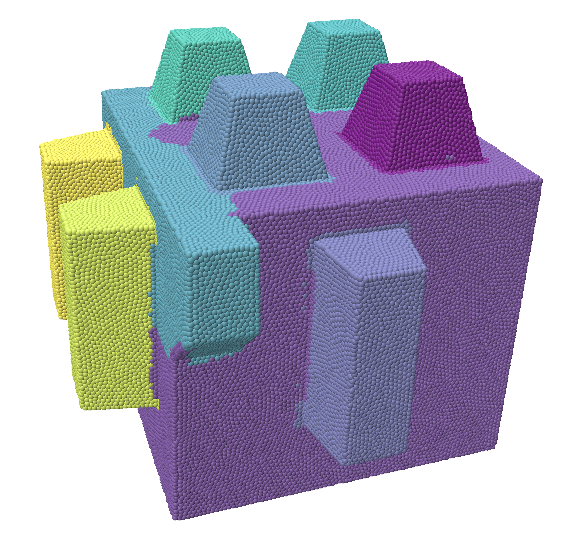}
	\caption{Results of intermediate steps of WCSEG. Left: Over-segmentation. Right: Results after merging.}
	\label{fig:wcseg_steps}
\end{figure}
As an alternative to LoS, we have also experimented with the Weakly Convex Segmentation (WCSEG) approach introduced in \cite{kaick2014shape}. 
\\
\textbf{Over-Segmentation.} At first, the input point-cloud is over-segmented into multiple small patches, obtained by using a region-growing approach that considers neighbors, in the $k$-NN sense, with "close" normal vectors (i.e. the angle between the two normal vectors is below some threshold). This step is illustrated in Fig.~\ref{fig:wcseg_steps}, left image. 
\\
\textbf{Neighbor Patch Merging.} These multiple small patches are then merged with a region-growing based approach, where two neighbor patches are merged if they are mutually visible (using a notion of visibility similar to the one described in Section \ref{ch:los_clustering}). 
\\
\textbf{Volumetric Merging.} The result is a segmentation of the input point-cloud into weakly convex components. The clustering can be further improved by merging components using a volumetric dissimilarity score between two components, computed using the Shape Diameter Function (SDF) value introduced in \cite{shapira_sdf08}. An illustration of this step is provided in Fig.~\ref{fig:wcseg_steps}, right image. 

\subsubsection{Structured Point-cloud Assignment}
The union of all per-cluster point-clouds may differ from the set of structured points $O_s$ depending on the clustering method used (LoS performs a sub sampling, while we found experimentally that WCSEG performs better with the full input point cloud).
In order to even these differences, points of the structured point-cloud $O_s$ are re-distributed among all clusters $C$ such that $\bigcup_{c \in C} O_c = O_s$. 
This is done using efficient nearest neighbor queries to find for each point $o \in O_s$ the cluster containing the closest point to $o$ among all points of all clusters.

\subsection{Polytope Generation}
Based on the weakly convex clustering step, an optimization method is used to find the best combination of planes to form polytopes for each cluster $c \in C$.
Please note that if the clustering is not perfect (in the sense that it doesn't correspond to a convex part), more than a single polytope can be necessary to fit the target shape. Thus, our optimization method is capable to generate a set of convex polytopes per cluster, rather than a single polytope. 

%\subsubsection{Plane Neighborhood Graph Extraction}
%For efficient plane selection in the optimization process, it is necessary to  efficiently store plane neighborhood relations. 
%This is done by extracting a plane neighborhood graph $G_N = (P, N)$ with planes $P$ as vertices and edges $N$ whenever two planes are neighbors as defined in Chapter \ref{ch:structuring}.   
%The edge set $E$ is generated by iterating through all planes $p \in P$, retrieving their associated points and for each point $o \in O_s$, looking up the neighbor planes $\{n_{p;o}\}$ via $f_n$ and adding edges $(p, n_{p;o})$ to $E$. 

\subsubsection{Target Volume Creation}
This step and the following ones (Sections~\ref{ch:ga} and \ref{ch:polyfilter}) are performed for each cluster. 
Based on the cluster point-cloud $O_{c}$, a signed distance grid $V_c: \mathbb{R}^3 \to \mathbb{R}$ that maps 3D query points to signed distance values is extracted. 
$V_c$ represents the target volume of the cluster and is generated in order to define a meaningful measure for geometric fitness of a candidate solution (set of polytopes) in the optimization process.
This is done by reconstructing a surface mesh using Poisson surface reconstruction \cite{kazhdan2006poisson} taking the cluster point-cloud $O_{c}$ as input.
Then, for each grid center point, its signed distance to the reconstructed surface mesh is retrieved and stored. 
%To get a signed distance value for an arbitrary query point, we take the value of the cell the point is located in and add the signed distance to the cell's center.
The signed distance value at an arbitrary query point is obtained by interpolation. 

\subsubsection{Optimization}\label{ch:ga}
In this step, subsets of planes associated with a cluster are combined to form convex polytopes and the polytope set that best fits the cluster's target volume $V_c$ is selected.
This is done by formulating the search process as a combinatorial optimization problem over all cluster planes $P_c$ and solving it via an Evolutionary Algorithm. 
The population is a set of polytope sets (and each polytope consists of a subset of cluster planes $P_c$).
\\
\textbf{Objective Function.} The objective function to be optimized assigns a score to each individual $I$ of the population. 
It reads
\begin{equation}
F(I,O_c,V_c)= \alpha \cdot F_{g}(I,O_c) + \beta \cdot F_{p}(I,V_c) - \gamma \cdot \frac{|I|}{n_{I\; max}},
\end{equation}
where $F_{g}(\cdot,\cdot)$ is a geometric term, $F_{p}(\cdot,\cdot)$ is a per-polytope geometric term and the last term penalizes the size of $I$ and is normalized by a user-controlled maximum value $n_{I\; max}$. 
$\alpha$,$\beta$ and $\gamma$ are user-defined weighting parameters (in our experiments we chose $\alpha=1$, $\beta=1$, $\gamma=0.1$).
The geometric term reads 
\begin{equation}\label{eq:fgeo}
F_{g}(I,O_c) = \frac{1}{|O_c|} \sum_{o \in O_c}
\begin{cases}
1, & \text{if } \min_{i \in I} \; |d_i(o)| < \epsilon \\
0, & \text{otherwise} 
\end{cases},
\end{equation}
where $\epsilon$ is a user-defined minimum value for $d_i(\cdot)$ which is the distance of a 3D point to the surface of polytope $i \in I$.
In order to prevent polytope sets from containing polytopes with large parts being located outside of the target volume $V_c$, a per-polytope geometric term is needed. 
It reads 
\begin{equation}\label{eq:per_prim_geo}
F_{p}(I,V_c) = \sum_{i \in I} \frac{1}{|H_i|} \sum_{h \in H_i}
\begin{cases}
1, & \text{if } V_c(o_h) < \epsilon\\
0, & \text{otherwise} 
\end{cases},
\end{equation}
where $H_i$ is the set of voxels representing the discretized volume of polytope $i$ and $\epsilon$ is a user-defined distance threshold. 
\\
\textbf{Initialization.} The initial population is filled with randomly generated polytope sets containing polytopes assembled as follows: 
At first, a plane is randomly selected from $P_c$. 
Then, a randomly selected number of neighbor planes are added from the neighborhood graph $G_N$.
In some cases, the plane normals of the the polytope's faces are not consistently oriented and need to be flipped. 
For computing the objective function later on, it is necessary to have a discretized volume representation (voxel grid) of the polytope.
For that purpose, the polytope's hull points are computed using the Double Description method \cite{Fukuda1995DoubleDM}.
%With a perfect convex clustering, the EA would only need to have a single polytope per individual. 
%However, the weak clustering approaches used for this work do not always result in a perfectly convex decomposition.
%Thus, an individual needs to contain more than one polytope.
\\
\textbf{Variation Operators.}
The Crossover operator exchanges randomly selected sequences of polytopes in two individuals. 
The Mutation operator modifies a single individual and has multiple modes: 
1) Alter the polytope set by randomly replacing polytopes with newly created random polytopes.
2) Add or remove random polytopes. 
3) Modify existing polytopes by adding randomly selected planes.
\\
\textbf{Elite Selection.}
After each iteration, the polytopes with the population-wide best per-polytope geometry scores are selected to form a new individual in the next iteration.
Experiments revealed that this procedure greatly improves convergence speed.
\\
\textbf{Termination.} The Evolutionary Algorithm terminates if either a certain maximum iteration limit is reached or if the score of the best solution candidate has not improved over a certain number of iterations.

\subsubsection{Polytope Filtering}\label{ch:polyfilter}
The polytopes generated per cluster with a geometry score lower than a particular threshold are removed from the result set. 
In addition, duplicates (polytopes using the same planes) are eliminated and polytopes that are fully contained in another polytope are removed as well. 
% Not needed 
%\subsection{Polytope Merging}
%The weak convex clustering might result in a too rigorous convex decomposition where merged neighboring polytopes would still be forming a convex shape. 
%In order to reduce the amount of polytopes, the following procedure is applied:
%\todo[inline]{describe the merging procedure.}

\section{\uppercase{Evaluation}}
\label{sec:evaluation}
We used seven models in our evaluation - all with different shape and complexity. 
See Table \ref{tab:pc_sizes} for the sizes of the corresponding point-clouds. 
\begin{table}[ht]
\caption{Point-cloud sizes for all models.}
\begin{center}
\begin{tabular}{| C{0.6cm} | C{0.6cm} | C{0.6cm} | C{0.6cm} | C{0.6cm} | C{0.6cm} | C{0.6cm} |}
\hline
\textbf{M$1$} & \textbf{M$2$} & \textbf{M$3$} & \textbf{M$4$} & \textbf{M$5$} & \textbf{M$6$} & \textbf{M$7$} \\ 
\hline
$50$k & $43$k & $30$k & $30$k & $25$k & $25$k & $30$k \\  
\hline
\end{tabular}
\label{tab:pc_sizes}
\end{center}
\end{table}

\subsection{Weak Convex Clustering}
\label{ch:exp_clustering}
\begin{figure*}[!h!t!bp]
    \centering
    %\begin{subfigure}[h]{0.2\textwidth}
    \begin{subfigure}[h]{0.20\textwidth+20pt\relax}
        \centering
            \makebox[20pt]{\raisebox{40pt}{\rotatebox[origin=c]{90}{LoS}}}%
            \includegraphics[width=\dimexpr\linewidth-20pt\relax]{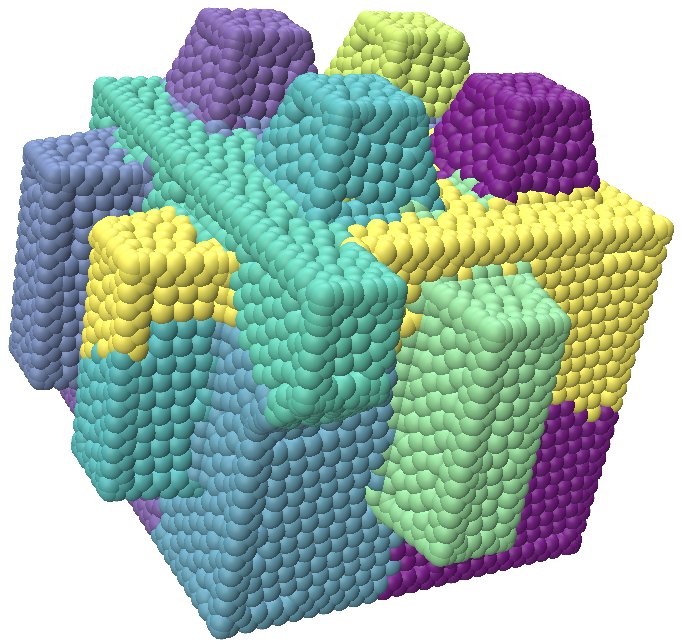}
    \end{subfigure}
    \begin{subfigure}[h]{0.2\textwidth}
        \centering
        \includegraphics[width=\textwidth]{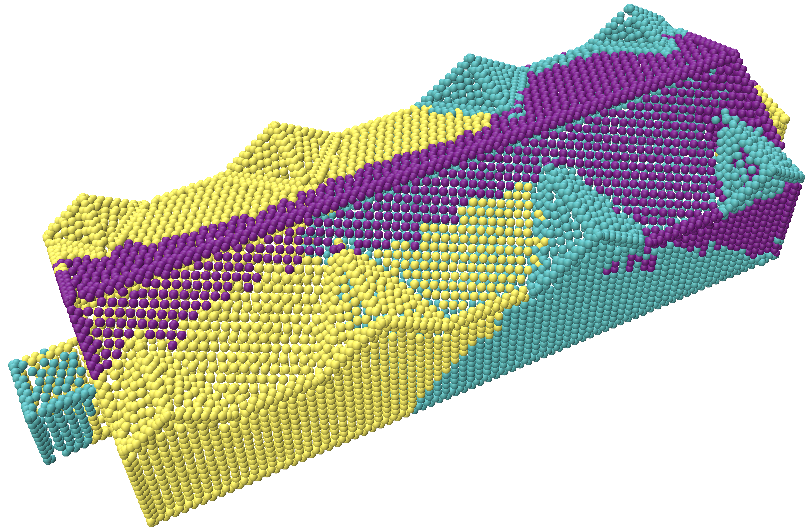}
    \end{subfigure}
    \begin{subfigure}[h]{0.2\textwidth}
        \centering
        \includegraphics[width=\textwidth]{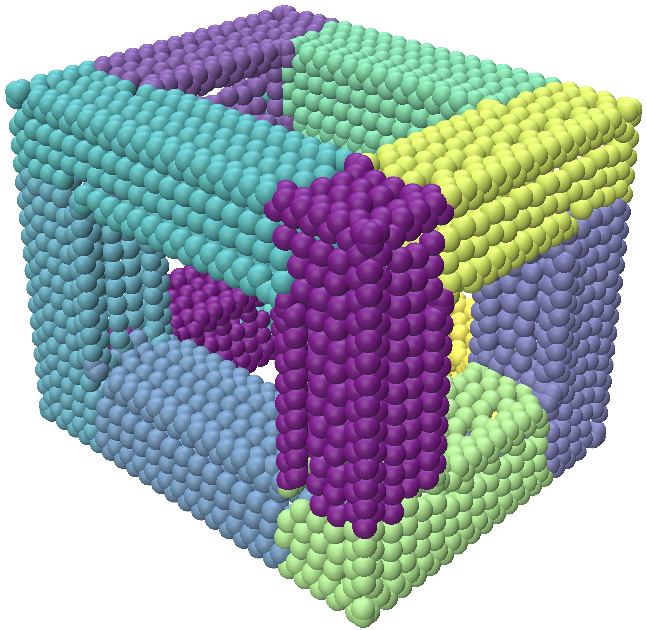}
    \end{subfigure}
     \begin{subfigure}[h]{0.15\textwidth}
        \centering
        \includegraphics[width=\textwidth]{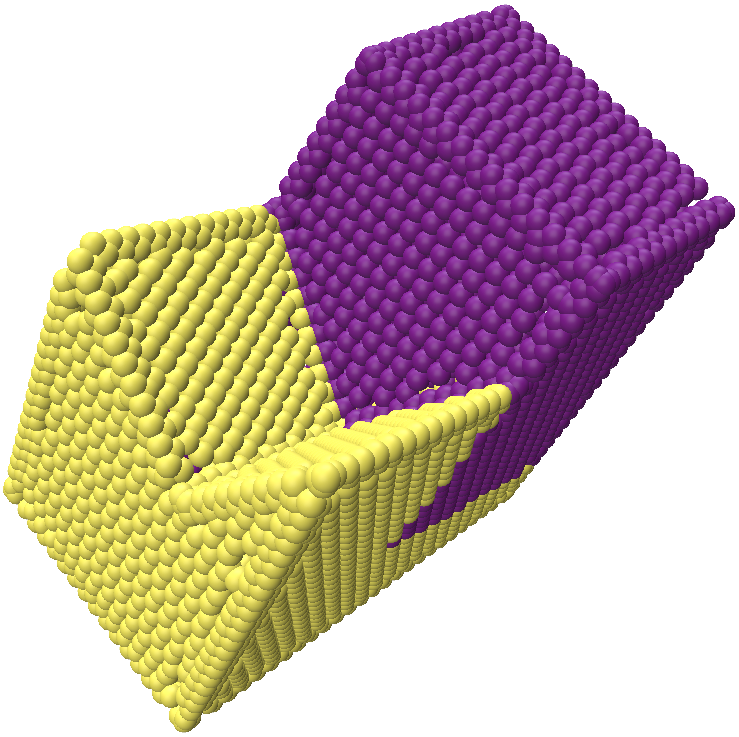}
    \end{subfigure}
    \\
    %\begin{subfigure}[h]{0.2\textwidth}
    \begin{subfigure}[h]{0.20\textwidth+20pt\relax}
        \centering
        \makebox[20pt]{\raisebox{40pt}{\rotatebox[origin=c]{90}{WCSEG}}}%
        \includegraphics[width=\dimexpr\linewidth-20pt\relax]{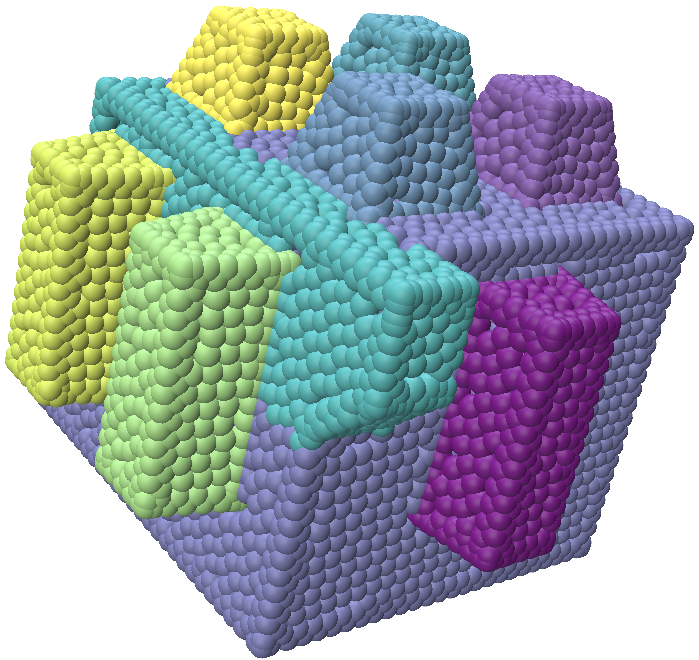}
    \end{subfigure}
    \begin{subfigure}[h]{0.2\textwidth}
        \centering
        \includegraphics[width=\textwidth]{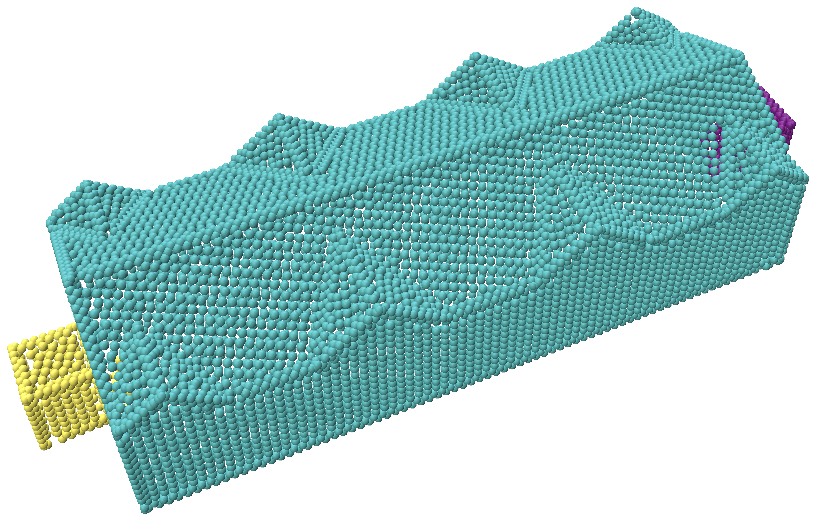}
    \end{subfigure}
    \begin{subfigure}[h]{0.2\textwidth}
        \centering
        \includegraphics[width=\textwidth]{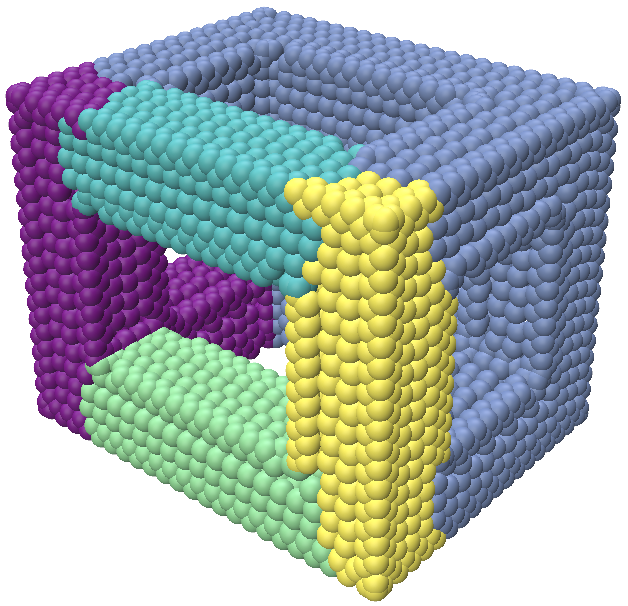}
    \end{subfigure}
    \begin{subfigure}[h]{0.15\textwidth}
        \centering
        \includegraphics[width=\textwidth]{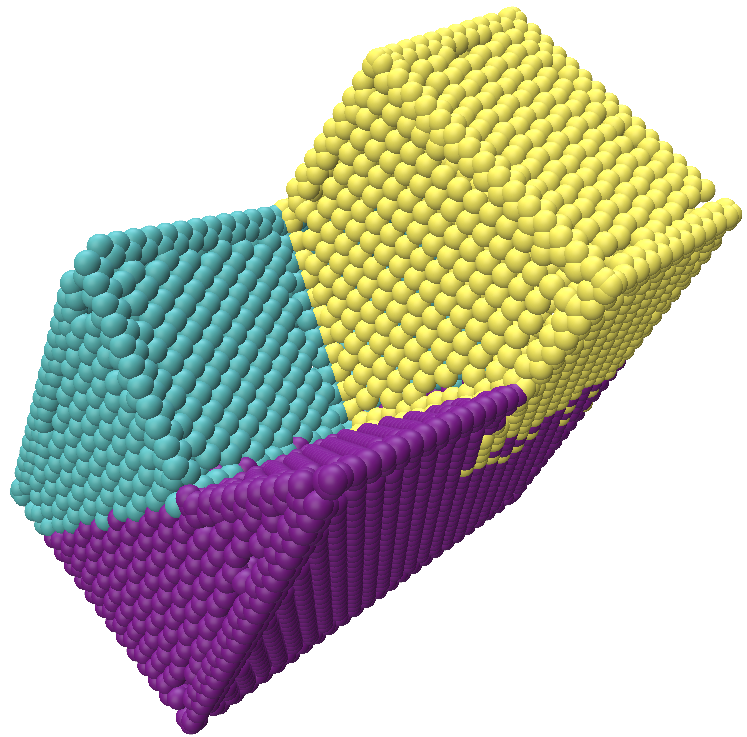}
    \end{subfigure}
    \caption{Clustering results for models M$1$, M$2$, M$4$ and M$7$ (left to right) for LoS (top row) and WCSEG (bottom row).} \label{fig:clustering_result}
\end{figure*}
As described in Section~\ref{ch:clustering}, we consider two approaches for weak convex clustering: a variant of LoS and WCSEG. 
In the following, we compare their results on the evaluation data-sets. 
The results obtained by these two approaches are summarized in Table \ref{tab:clustering_results}. In this table, the term "Perfect" refers to our qualitative opinion of what the correct number of convex clusters should be for a given model.
\\
As shown in Table \ref{tab:clustering_results}, the result quality depends on the input model. 
\\
LoS has perfect results for models M$3$ and M$4$ where WCSEG demonstrates a lower quality result (for M$4$, see Fig.~\ref{fig:clustering_result}, third column).
\\
On the other hand, WCSEG gives a perfect clustering result for model M$1$ whereas LoS is not able to handle the geometric details (see Fig.~\ref{fig:clustering_result}, first column). 
The same is true for model M$7$ where WCSEG performs well except for a few incorrectly classified points, while LoS is not able to find all three clusters (see Fig.~\ref{fig:clustering_result}, last column).
\\
Model M$2$ consists of an almost convex main part and two attached cuboids. 
Both approaches fail to classify the smaller details of the main part as independent convex clusters (see Fig.~\ref{fig:clustering_result}, second column). 
\\
WCSEG is able to find the 'almost' convex clusters of M$3$ whereas LoS fails to find meaningful convex clusters. 
\\
If all polytopes are spatially separated like in model M$6$, both LoS and WCSEG find the correct clusters.
\begin{table}[ht]
\caption{Number of clusters for both clustering approaches and all models. The "-" symbols indicate the amount of incorrect cluster assignments.}
\begin{center}
\begin{tabular}{| C{1cm} | C{1.75cm} | C{1.75cm} | C{1.25cm} |}
\hline
\textbf{Model} & \textbf{LoS} & \textbf{WCSEG} & \textbf{Perfect} \\ 
\hline
M$1$ & $10$ (- -) & $11$ & $11$ \\  
\hline
M$2$ & $3$ (- -) & $3$ & $11$ \\  
\hline
M$3$ & $12$ & $12$ (-) & $12$ \\  
\hline
M$4$ & $12$ & $5$ & $12$ \\  
\hline
M$5$ & $1$ & $2$ & $2$ \\  
\hline
M$6$ & $2$ & $2$ & $2$ \\  
\hline
M$7$ & $2$ & $3$ (-) & $3$ \\  
\hline
\end{tabular}
\label{tab:clustering_results}
\end{center}
\end{table}

\subsection{Pipeline Results}
\begin{figure*}[!h!t!bp]
\centering
\begin{subfigure}[t]{0.15\textwidth+20pt\relax}
    \makebox[20pt]{\raisebox{38pt}{\rotatebox[origin=c]{90}{M1}}}%
    \includegraphics[width=\dimexpr\linewidth-20pt\relax]
    {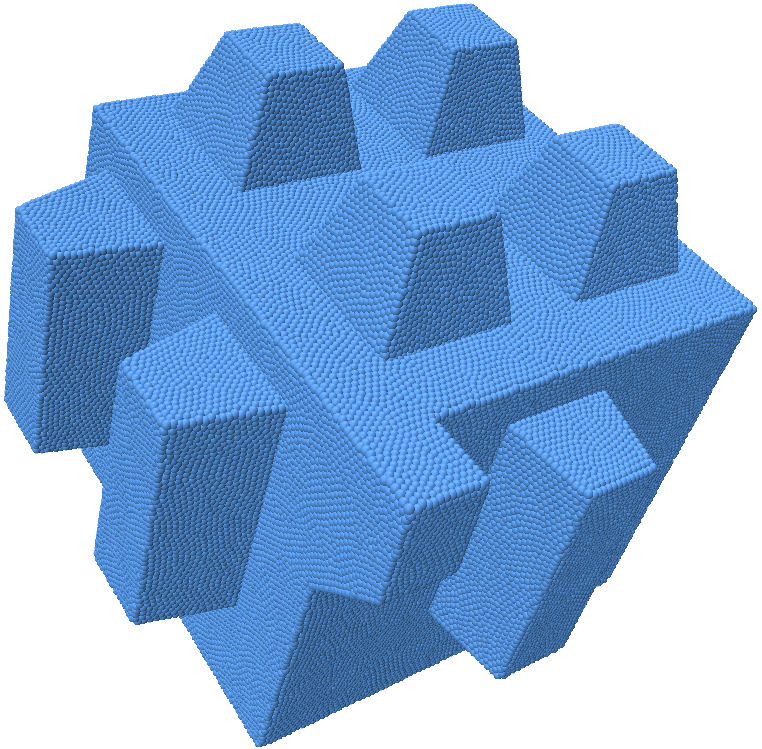}
    \makebox[20pt]{\raisebox{38pt}{\rotatebox[origin=c]{90}{M2}}}%
    \includegraphics[width=\dimexpr\linewidth-20pt\relax]
    {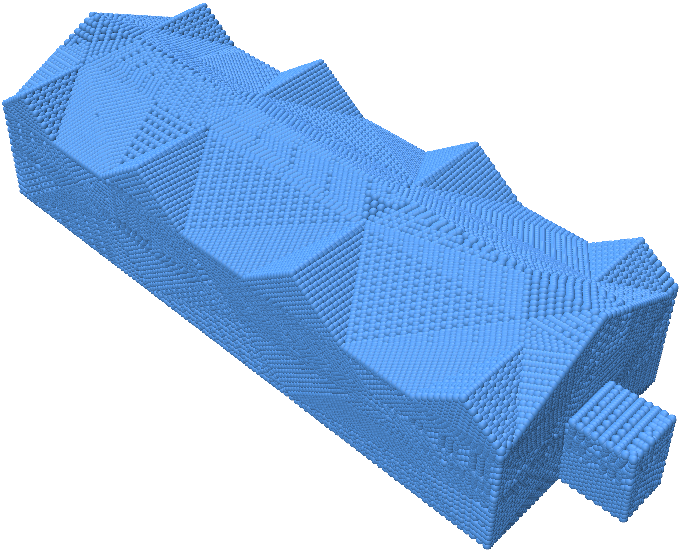}
    \makebox[20pt]{\raisebox{38pt}{\rotatebox[origin=c]{90}{M3}}}%
    \includegraphics[width=\dimexpr\linewidth-20pt\relax]
    {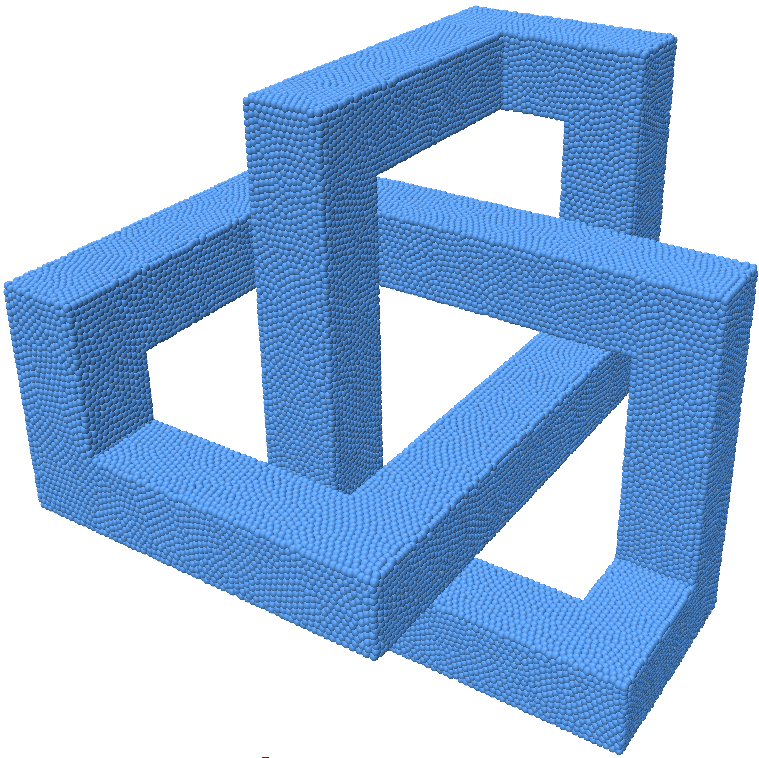}
    \makebox[20pt]{\raisebox{38pt}{\rotatebox[origin=c]{90}{M4}}}%
    \includegraphics[width=\dimexpr\linewidth-20pt\relax]
     {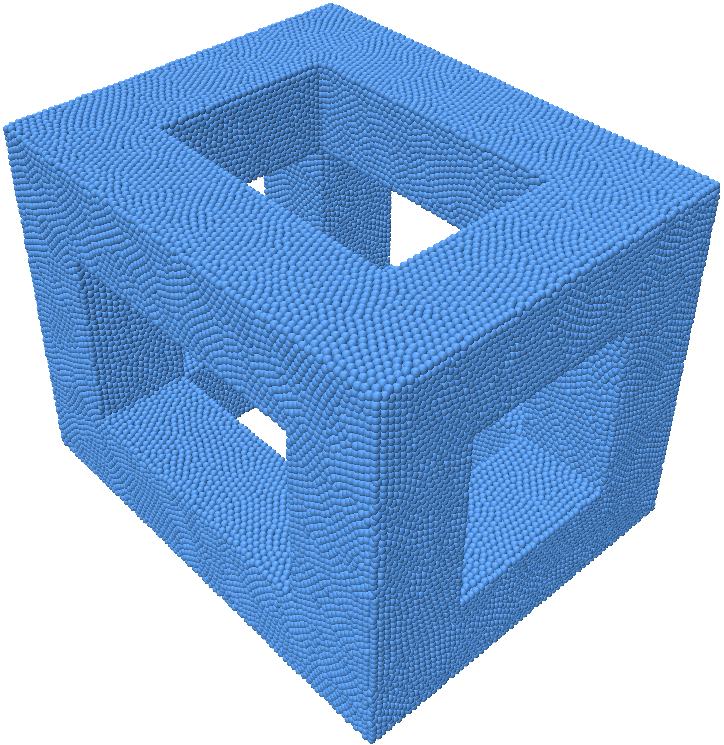}
    \makebox[20pt]{\raisebox{38pt}{\rotatebox[origin=c]{90}{M5}}}%
    \includegraphics[width=\dimexpr\linewidth-20pt\relax]
    {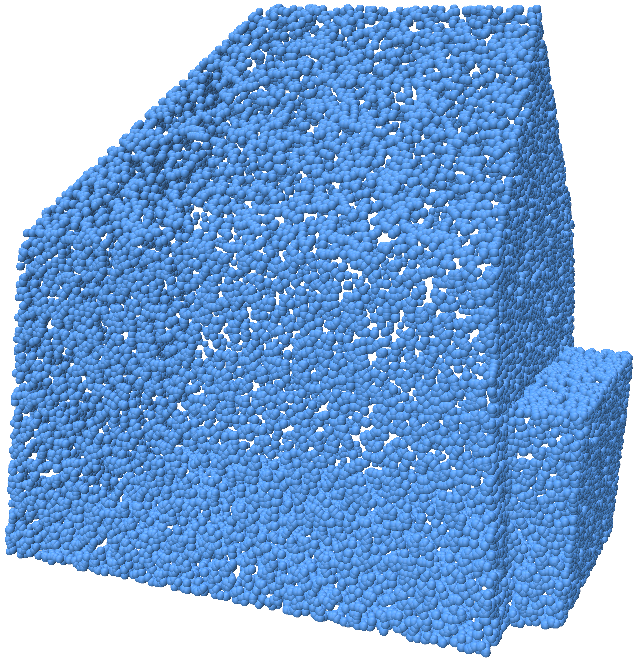}
    \makebox[20pt]{\raisebox{38pt}{\rotatebox[origin=c]{90}{M6}}}%
    \includegraphics[width=\dimexpr\linewidth-20pt\relax]
    {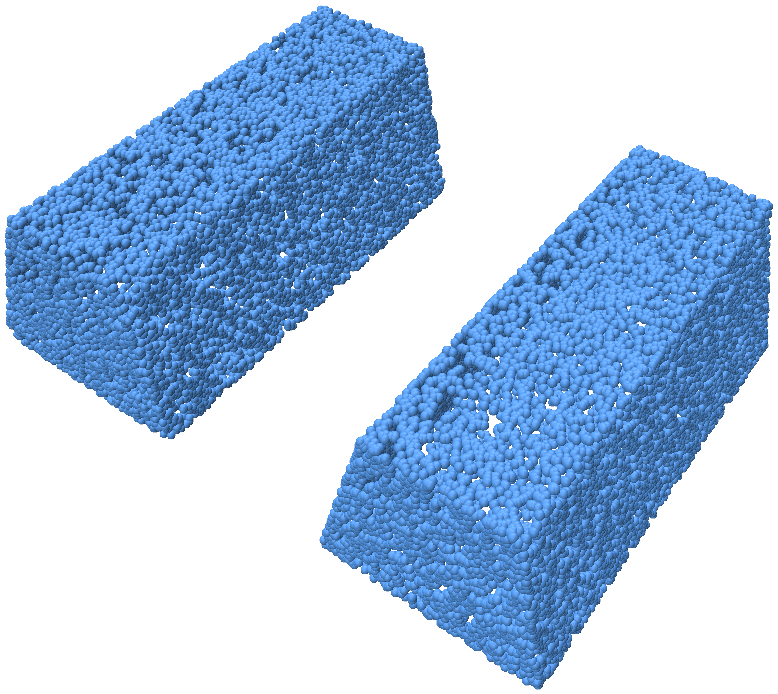}
    \makebox[20pt]{\raisebox{38pt}{\rotatebox[origin=c]{90}{M7}}}%
    \includegraphics[width=\dimexpr\linewidth-20pt\relax]
    {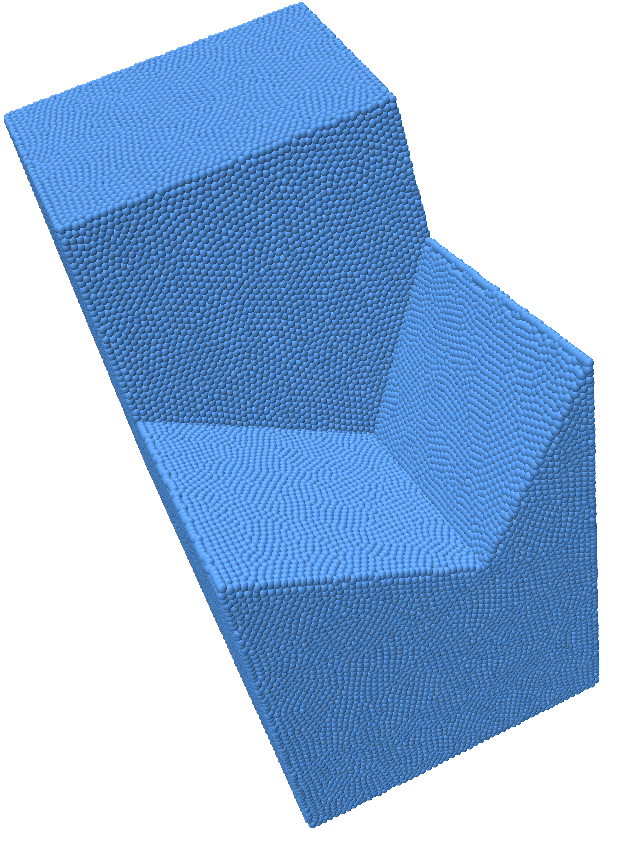}
    \caption{Input point-cloud.}\label{fig:input_pc}
\end{subfigure}\hfill
\begin{subfigure}[t]{0.15\textwidth}
    \includegraphics[width=\textwidth]
    {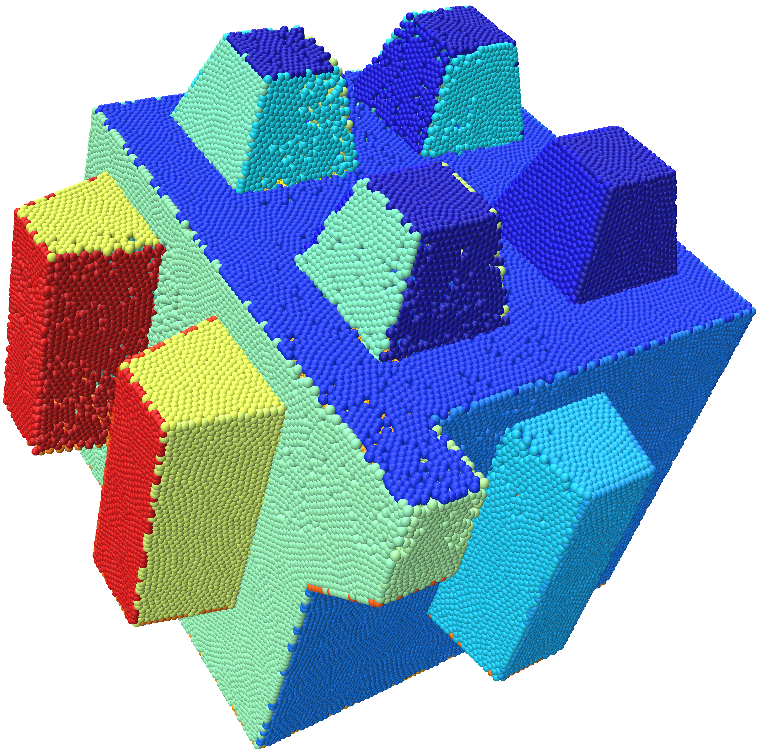}
    \includegraphics[width=\textwidth]
    {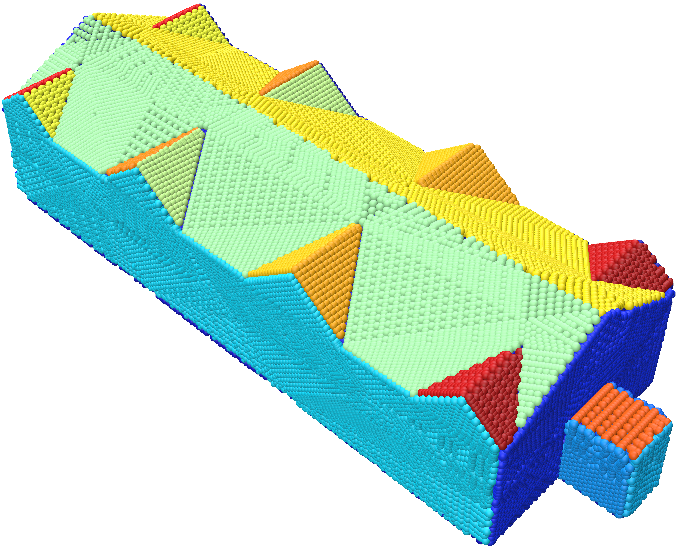}
    \includegraphics[width=\textwidth]
    {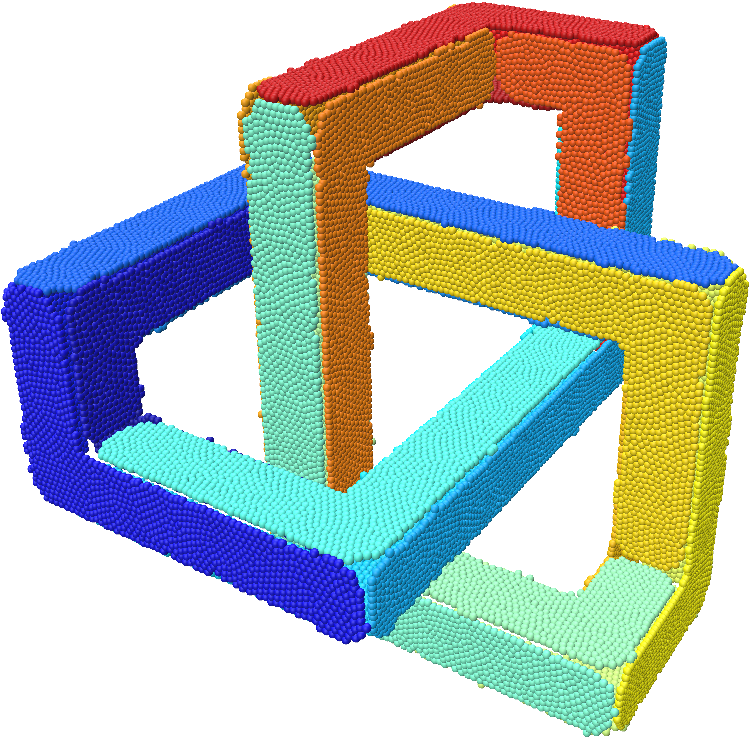}
    \includegraphics[width=\textwidth]
    {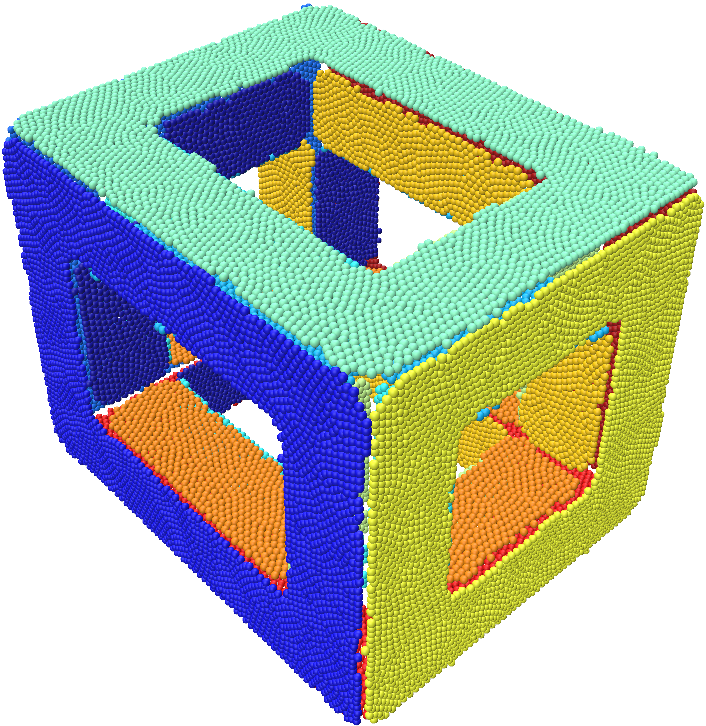}
    \includegraphics[width=\textwidth]
    {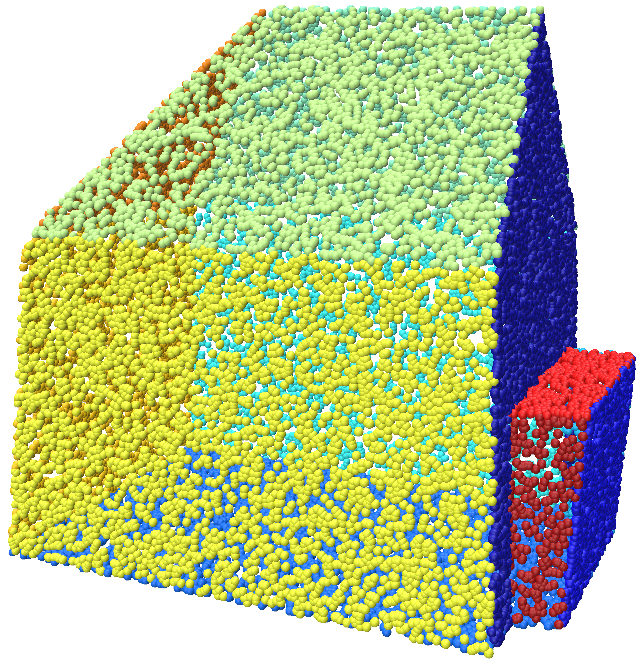}
     \includegraphics[width=\textwidth]
    {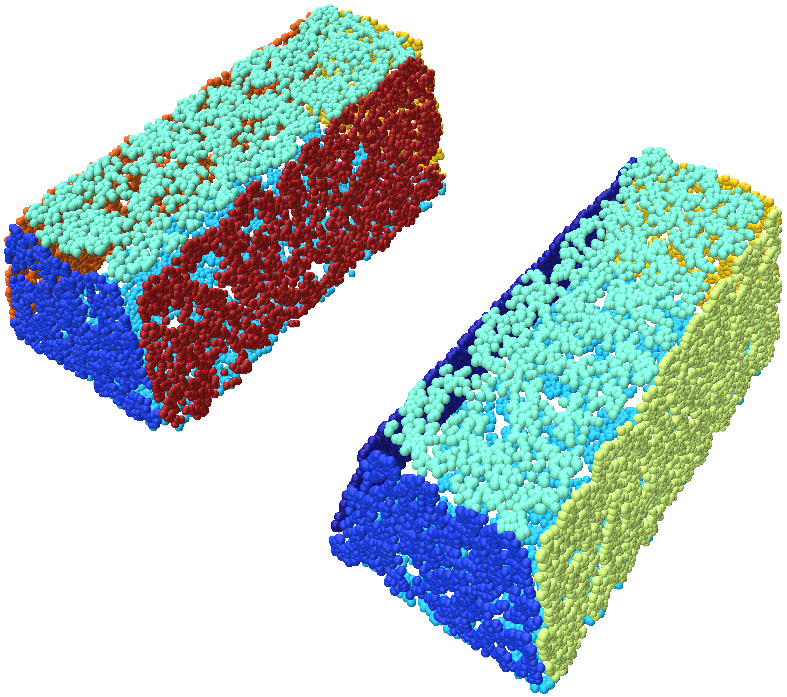}
     \includegraphics[width=\textwidth]
    {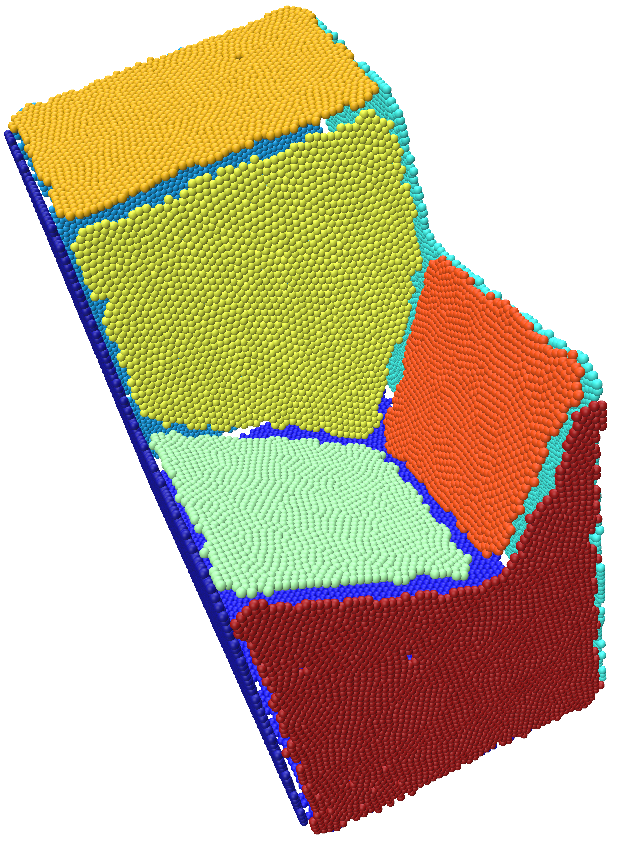}
    \caption{Points corresponding to fitted planes.}\label{fig:fit_planes}
\end{subfigure}\hfill
\begin{subfigure}[t]{0.15\textwidth}
    \includegraphics[width=\textwidth]  
    {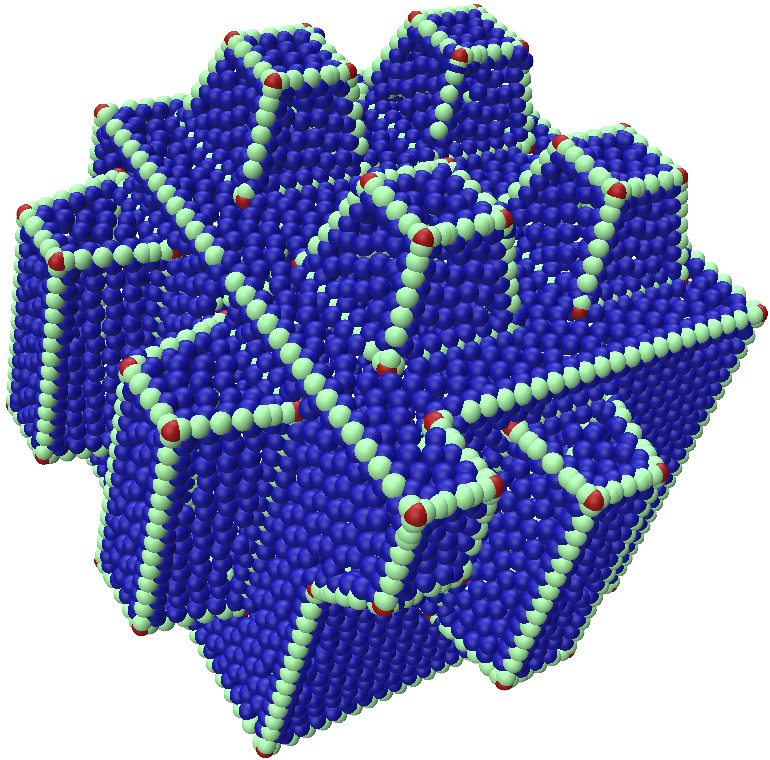}
    \includegraphics[width=\textwidth]
    {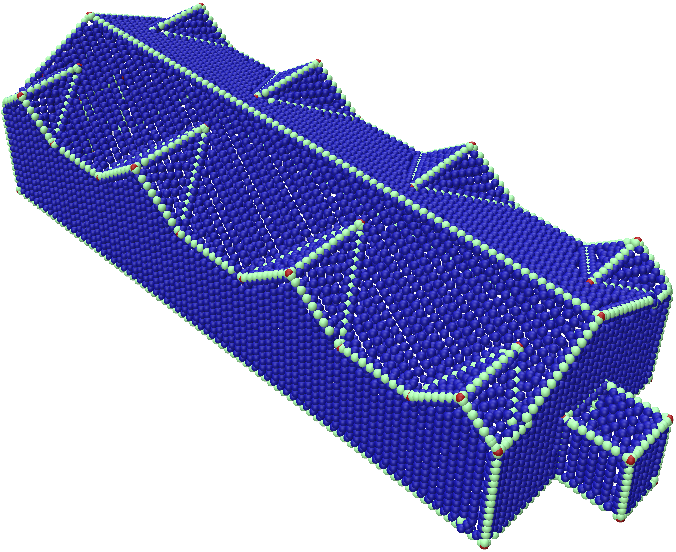}
    \includegraphics[width=\textwidth]
    {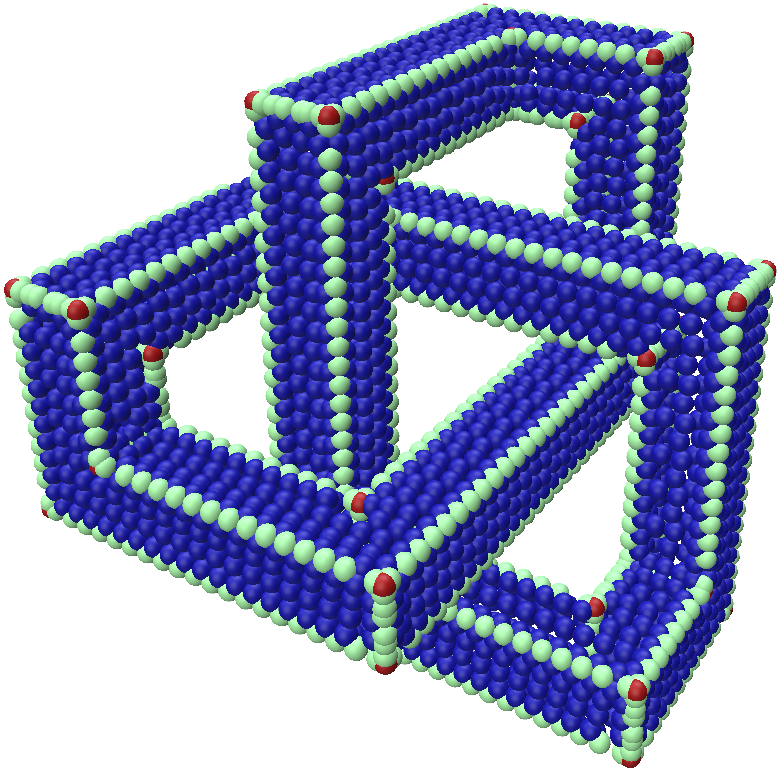}
    \includegraphics[width=\textwidth]
    {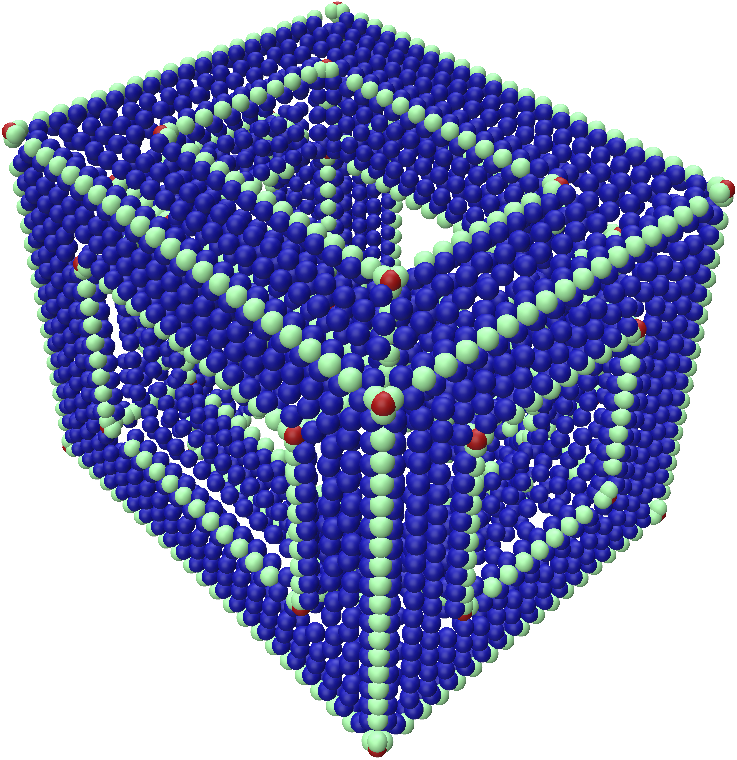}
    \includegraphics[width=\textwidth]
    {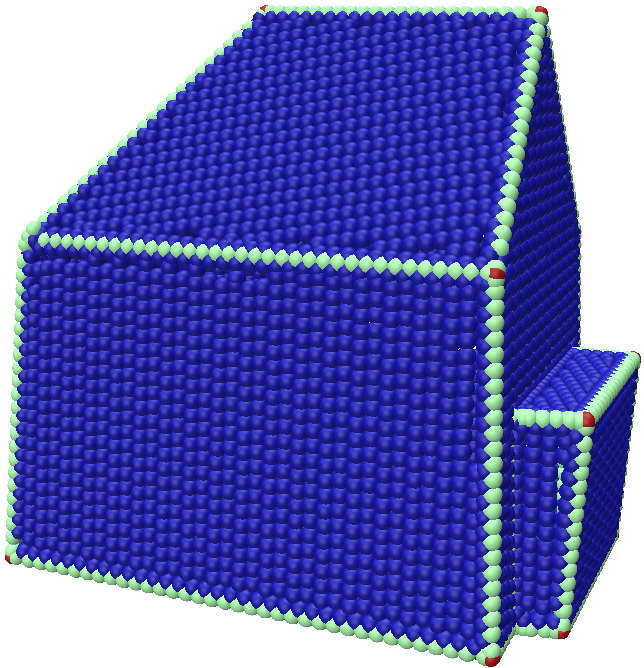}
    \includegraphics[width=\textwidth]
    {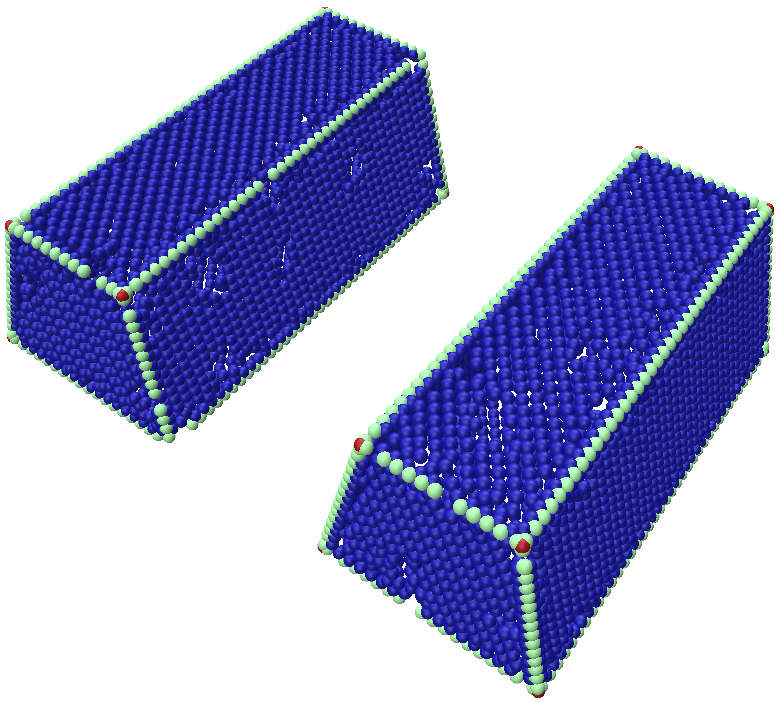}
    \includegraphics[width=\textwidth]
    {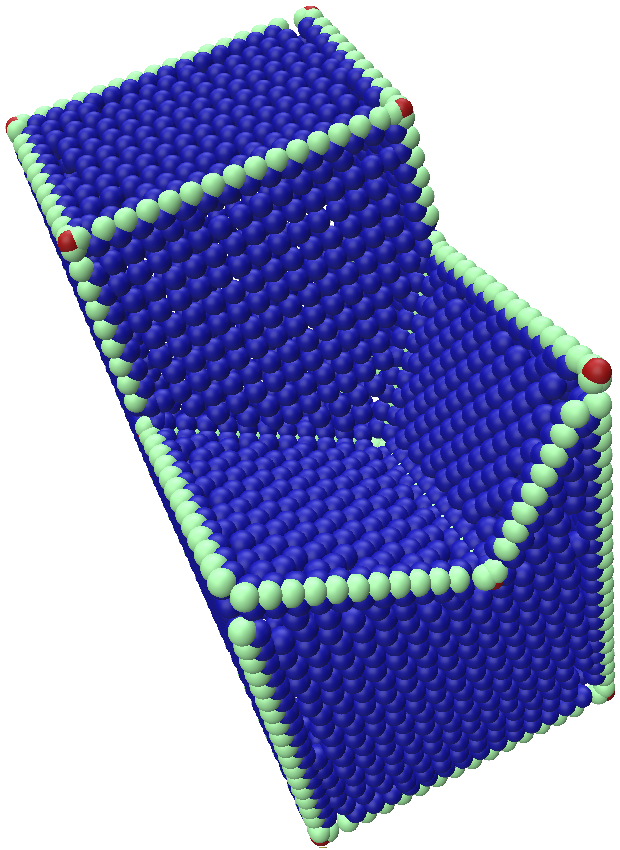}
    \caption{Structured point-cloud.}\label{fig:structured}
\end{subfigure}\hfill
\begin{subfigure}[t]{0.15\textwidth}
    \includegraphics[width=\textwidth]  
     {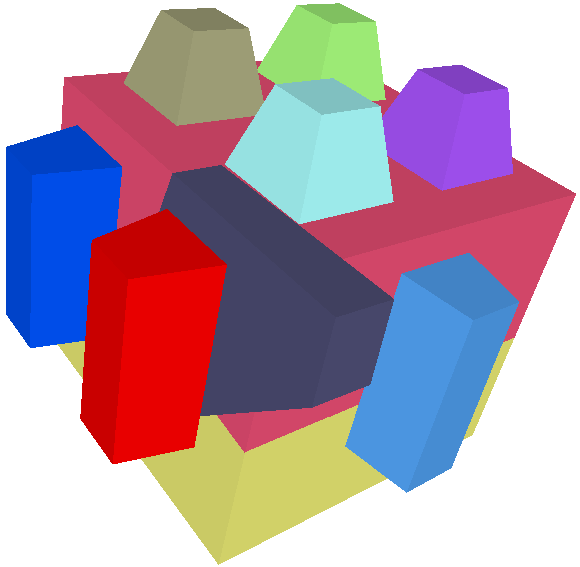}
    \includegraphics[width=\textwidth]
    {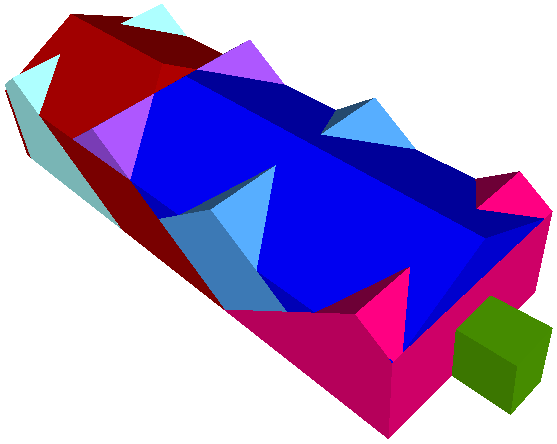}
    \includegraphics[width=\textwidth]
    {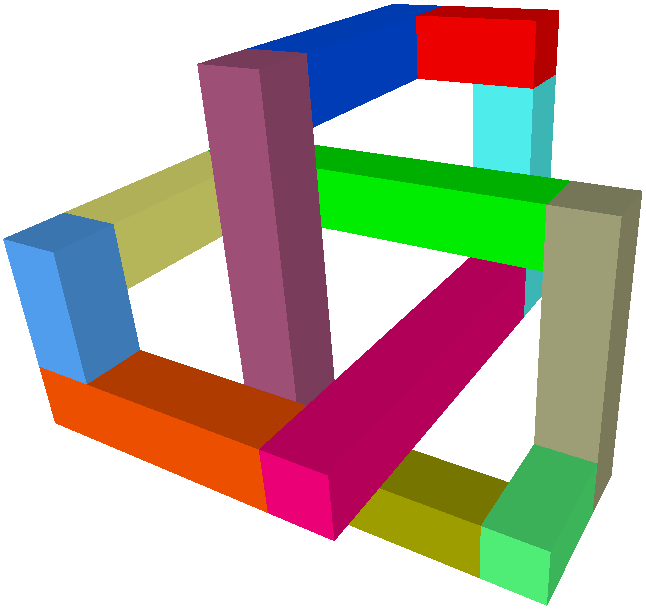}
    \includegraphics[width=\textwidth]
    {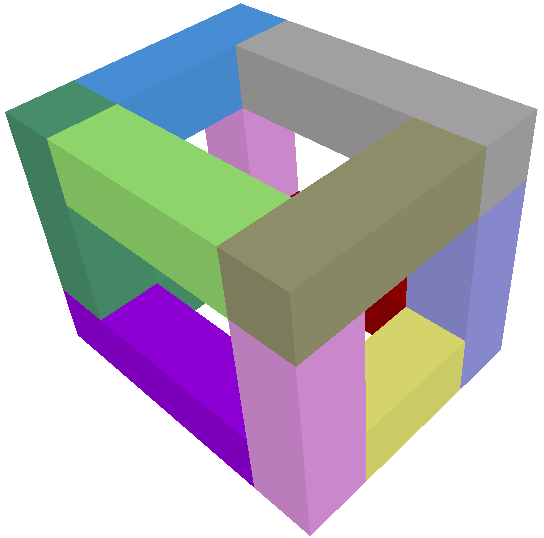}
    \includegraphics[width=\textwidth]
    {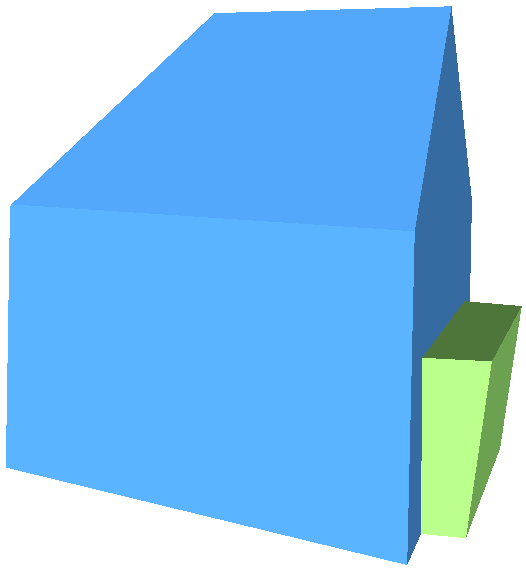}
    \includegraphics[width=\textwidth]
    {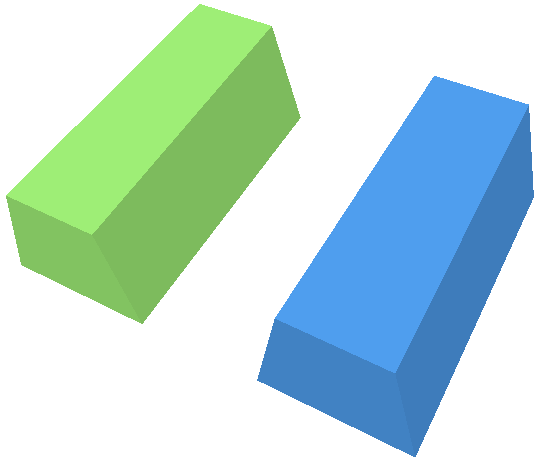}
    \includegraphics[width=\textwidth]
    {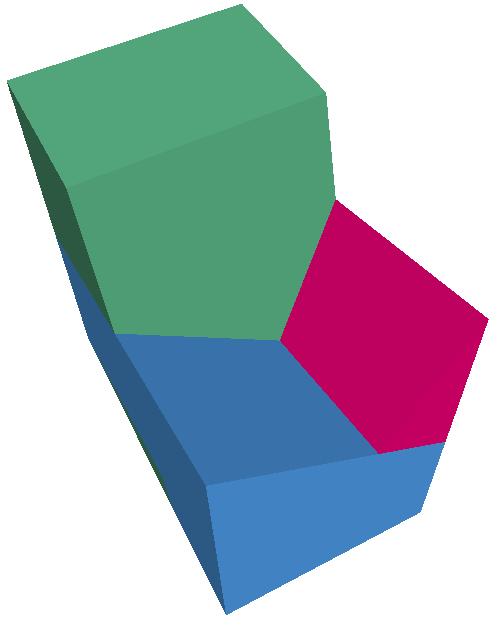}
    \vspace*{3mm}\caption{Generated convex polytopes.}\label{fig:results}
\end{subfigure}
\vspace*{5mm}\caption{Pipeline results for all models.}\label{fig:all_results}
\end{figure*}

Fig.~\ref{fig:all_results} shows the input point-clouds for our test models (Fig.~\ref{fig:input_pc}) as well as the visual results of the different pipeline steps: Plane Extraction (Section~\ref{ch:plane_extraction}) in Fig.~\ref{fig:fit_planes}, Point-cloud Structuring (Section~\ref{ch:structuring}) in Fig.~\ref{fig:structured}, and Polytope Generation (Section~\ref{ch:ga}) in Fig.~\ref{fig:results}. 
For the results of point-cloud structuring (Fig.~\ref{fig:structured}), the blue color indicates planar points, the green color crease points and the brown color corner points. 
For each model, the best possible clustering approach was used (See Sections~\ref{ch:clustering} and \ref{ch:exp_clustering}).
All models are correctly reconstructed. 
Only M$2$ has a small part of the roof incorrectly recovered (dark red polytope), but taking into account the low clustering quality for this model (see Fig.~\ref{fig:clustering_result}, second column), the resulting polytope collection is actually good.
The result of M$6$ shows that the pipeline can also correctly handle disconnected input shapes. 

\begin{figure*}[!h!t!bp]
\centering
\begin{subfigure}[h]{0.3\textwidth}
	\includegraphics[width=\linewidth]{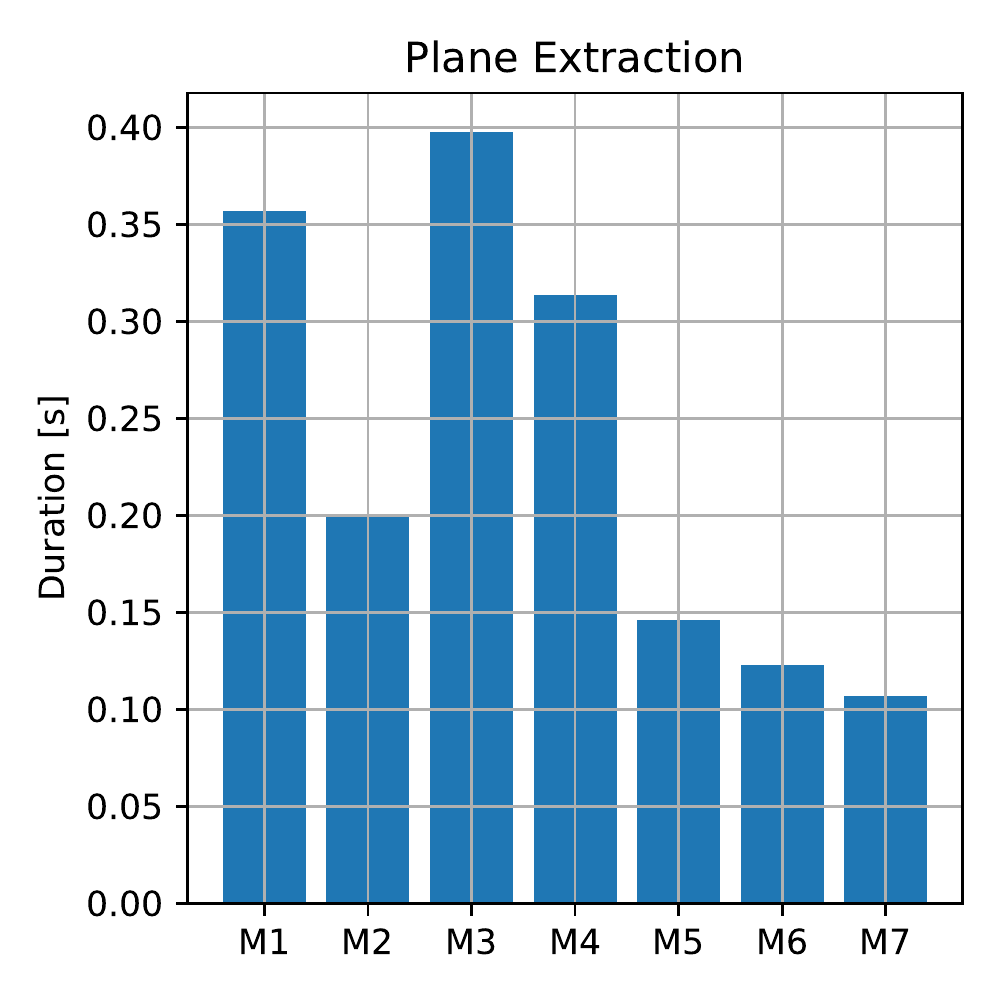}
	\caption{Plane Extraction.}
	\label{fig:plane_extract_timings}
\end{subfigure}
\begin{subfigure}[h]{0.3\textwidth}
	\includegraphics[width=\linewidth]{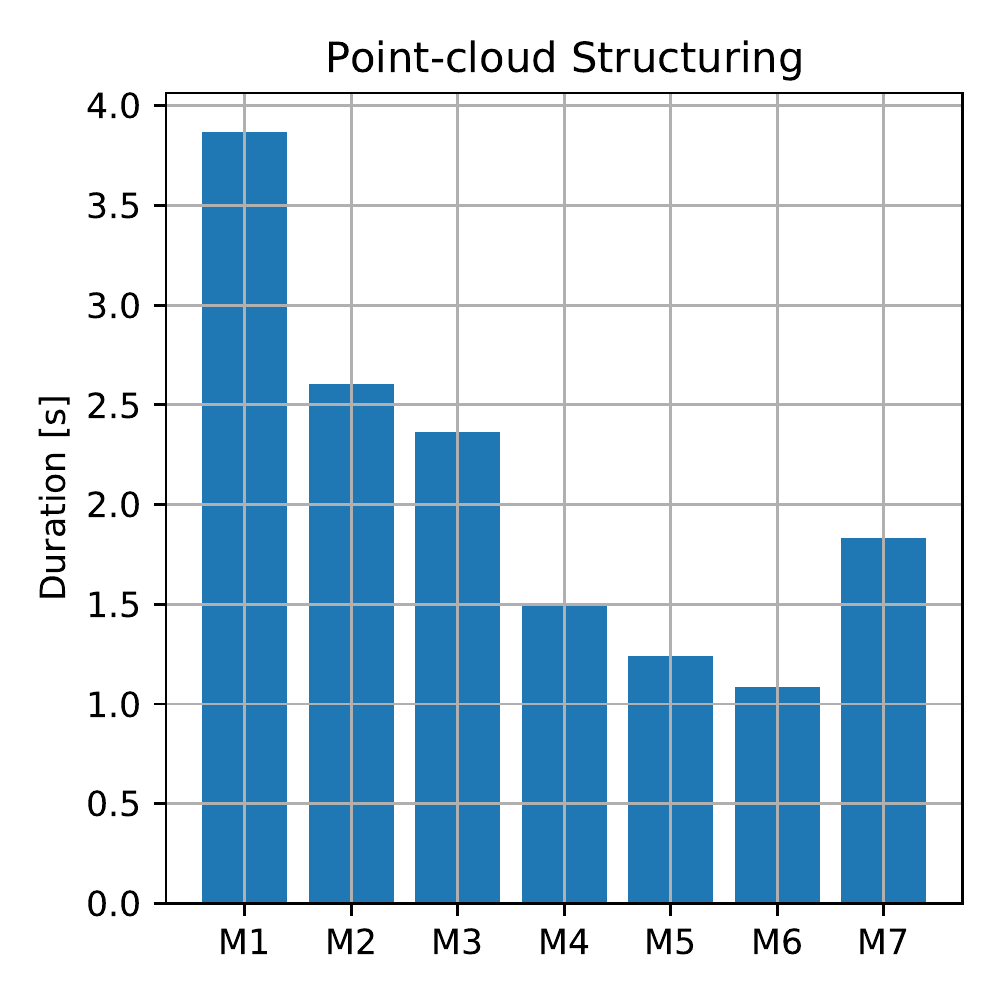}
	\caption{Point-cloud Structuring.}
	\label{fig:pc_struct_timings}
\end{subfigure}
\begin{subfigure}[h]{0.3\textwidth}
	\includegraphics[width=\linewidth]{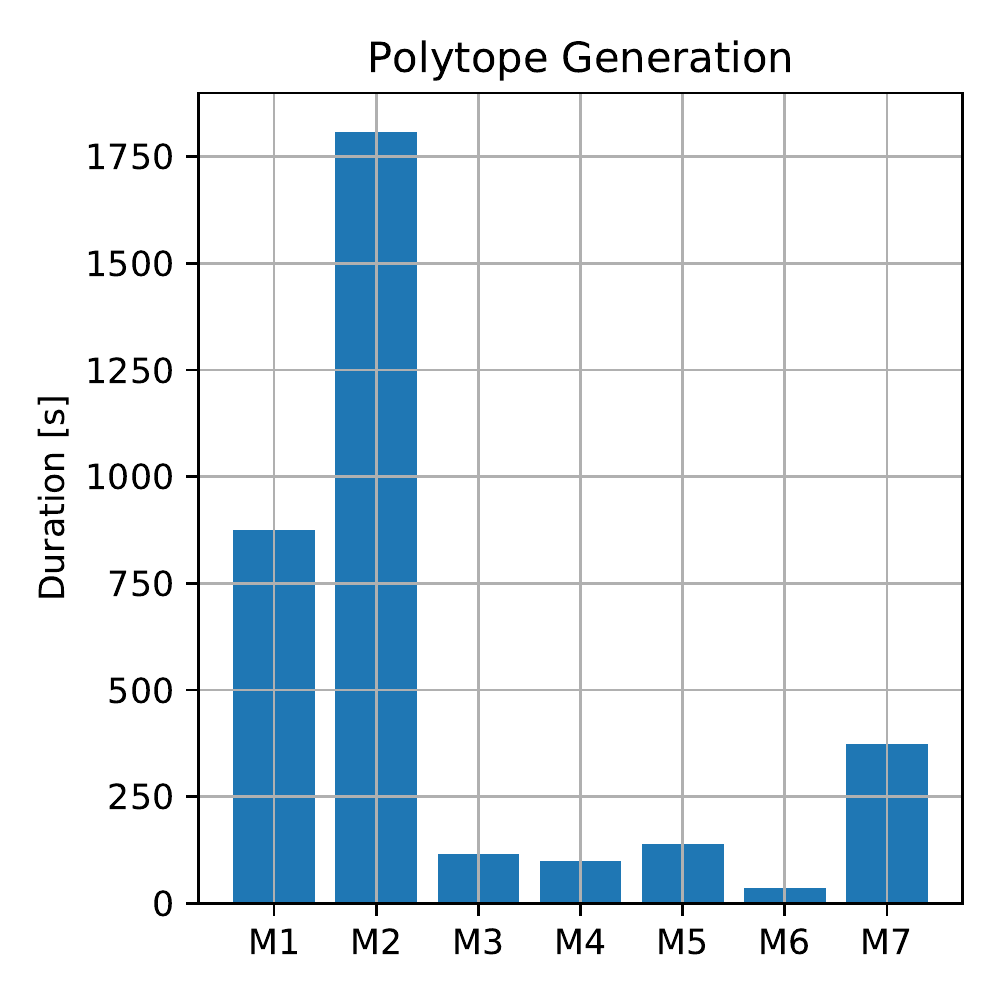}
	\caption{Polytope Generation.}
	\label{fig:poly_gen_timings}
\end{subfigure}
\vspace*{5mm}\caption{Timings for the main pipeline steps.}
\end{figure*}

The wall-clock times for the pipeline steps Plane Extraction, Point-cloud Structuring and Polytope Generation were measured on a Notebook running a $2.4$GHz dual-core CPU with $16$GB of RAM and are shown in Fig.~\ref{fig:plane_extract_timings}, \ref{fig:pc_struct_timings} and \ref{fig:poly_gen_timings}.
For all models, the Polytope Generation step takes the most time (see Fig.~\ref{fig:poly_gen_timings}), due to the use of an Evolutionary Algorithm (Section~\ref{ch:ga}).  
The EA takes the most time for M$2$ since it has to generate polytopes for all the roof details due to the coarse clustering (only $3$ clusters found). 
Interestingly, EA timings for M$7$ are relatively high. 
This can be explained as follows: 
The Point-cloud Structuring step produces an incomplete neighborhood graph.
Thus, the EA needs to consider all model planes in order to obtain a perfect reconstruction which increases the search space significantly.
By comparison, the timings for the plane extraction step (see Fig.~\ref{fig:plane_extract_timings}) and the point-cloud structuring step (see Fig.~\ref{fig:pc_struct_timings}) are negligible. Plane Extraction takes under $1$s for all models, whereas Point-cloud Structuring takes between $1.1$s (M$6$) and $3.9$s (M$1$).

\section{\uppercase{Conclusion}}
\label{sec:conclusion}
This paper introduces a pipeline for the reconstruction of a solid object as a collection (union) of convex polytopes from a point-cloud.
Two different approaches for partitioning the point-cloud in weak convex clusters are proposed. 
For each cluster, an EA is used to find the optimal set of polytopes from the extracted planes associated to that cluster.
%The approach works well on the evaluation data-sets. 
As future work, we are considering different possible directions: One is to further increase the scalability and robustness of the convex clustering approaches in order to deal with even larger and more complex point-clouds. Another direction of future work is to combine this approach with general approaches for CSG recovery \cite{fayolle2016evolutionary,friedrich_gecco2019,du2018inversecsg}. 

\section*{\uppercase{Acknowledgments}}
We would like to thank the anonymous reviewers for their valuable comments and suggestions.

\bibliographystyle{apalike}
%{\small
%\bibliography{bib}}
\bibliography{bib}

\begin{thebibliography}{}

\bibitem[Asafi et~al., 2013]{asafi2013weak}
Asafi, S., Goren, A., and Cohen-Or, D. (2013).
\newblock Weak convex decomposition by lines-of-sight.
\newblock {\em Computer graphics forum}, 32(5):23--31.

\bibitem[Benk{\'o} and V{\'a}rady, 2004]{BV04}
Benk{\'o}, P. and V{\'a}rady, T. (2004).
\newblock {Segmentation methods for smooth point regions of conventional
  engineering objects}.
\newblock {\em Computer-Aided Design}, 36(6):511--523.

\bibitem[Chen et~al., 2019]{chen2019bsp}
Chen, Z., Tagliasacchi, A., and Zhang, H. (2019).
\newblock Bsp-net: Generating compact meshes via binary space partitioning.
\newblock {\em arXiv preprint arXiv:1911.06971}.

\bibitem[Coumans and Bai, 2019]{coumans2019}
Coumans, E. and Bai, Y. (2016--2019).
\newblock Pybullet, a python module for physics simulation for games, robotics
  and machine learning.
\newblock \url{http://pybullet.org}.

\bibitem[Deng et~al., 2019]{deng2019cvxnets}
Deng, B., Genova, K., Yazdani, S., Bouaziz, S., Hinton, G., and Tagliasacchi,
  A. (2019).
\newblock Cvxnets: Learnable convex decomposition.
\newblock {\em arXiv preprint arXiv:1909.05736}.

\bibitem[Du et~al., 2018]{du2018inversecsg}
Du, T., Inala, J.~P., Pu, Y., Spielberg, A., Schulz, A., Rus, D., Solar-Lezama,
  A., and Matusik, W. (2018).
\newblock Inversecsg: Automatic conversion of 3d models to csg trees.
\newblock {\em ACM Trans. Graph.}, 37(6):1--16.

\bibitem[Edelsbrunner and M\"{u}cke, 1994]{Edelsbrunner1994}
Edelsbrunner, H. and M\"{u}cke, E.~P. (1994).
\newblock Three-dimensional alpha shapes.
\newblock {\em Transactions on Graphics}, 13(1):43–72.

\bibitem[Ester et~al., 1996]{Ester1996DBSCAN}
Ester, M., Kriegel, H.-P., Sander, J., and Xu, X. (1996).
\newblock A density-based algorithm for discovering clusters in large spatial
  databases with noise.
\newblock In {\em Proceedings of the Second International Conference on
  Knowledge Discovery and Data Mining}, KDD'96, pages 226--231. AAAI Press.

\bibitem[Fayolle and Pasko, 2016]{fayolle2016evolutionary}
Fayolle, P.-A. and Pasko, A. (2016).
\newblock An evolutionary approach to the extraction of object construction
  trees from 3d point clouds.
\newblock {\em Computer-Aided Design}, 74:1--17.

\bibitem[Fischler and Bolles, 1981]{fischler1981random}
Fischler, M.~A. and Bolles, R.~C. (1981).
\newblock Random sample consensus: a paradigm for model fitting with
  applications to image analysis and automated cartography.
\newblock {\em Communications of the ACM}, 24(6):381--395.

\bibitem[Friedrich et~al., 2019]{friedrich_gecco2019}
Friedrich, M., Fayolle, P.-A., Gabor, T., and Linnhoff-Popien, C. (2019).
\newblock Optimizing evolutionary {CSG} tree extraction.
\newblock In {\em Proceedings of the Genetic and Evolutionary Computation
  Conference}, GECCO ’19, page 1183–1191.

\bibitem[Friedrich et~al., 2020]{friedrich2020hybrid}
Friedrich, M., Illium, S., Fayolle, P.-A., and Linnhoff-Popien, C. (2020).
\newblock A hybrid approach for segmenting and fitting solid primitives to 3d
  point clouds.
\newblock In {\em Proceedings of the 15th International Conference on Computer
  Graphics Theory and Applications (GRAPP)}, volume~1, pages 38--48.

\bibitem[Fukuda and Prodon, 1996]{Fukuda1995DoubleDM}
Fukuda, K. and Prodon, A. (1996).
\newblock Double description method revisited.
\newblock In {\em Combinatorics and Computer Science}, pages 91--111. Springer
  Berlin Heidelberg.

\bibitem[Gilbert et~al., 1988]{Gilbert_jra88}
Gilbert, E.~G., Johnson, D.~W., and Keerthi, S.~S. (1988).
\newblock A fast procedure for computing the distance between complex objects
  in three-dimensional space.
\newblock {\em IEEE Journal on Robotics and Automation}, 4(2):193--203.

\bibitem[Kaick et~al., 2014]{kaick2014shape}
Kaick, O.~V., Fish, N., Kleiman, Y., Asafi, S., and Cohen-Or, D. (2014).
\newblock Shape segmentation by approximate convexity analysis.
\newblock {\em ACM Transactions on Graphics (TOG)}, 34(1):1--11.

\bibitem[Kaiser et~al., 2019]{Kaiser:2019:GeoPrimFitSurvey}
Kaiser, A., Zepeda, J. A.~Y., and Boubekeur, T. (2019).
\newblock A survey of simple geometric primitives detection methods for
  captured 3d data.
\newblock {\em Computer Graphics Forum}, 38(1):167--196.

\bibitem[Kazhdan et~al., 2006]{kazhdan2006poisson}
Kazhdan, M., Bolitho, M., and Hoppe, H. (2006).
\newblock Poisson surface reconstruction.
\newblock In {\em Proceedings of the fourth Eurographics symposium on Geometry
  processing}, volume~7 of {\em SGP '06}, page 61–70. Eurographics
  Association.

\bibitem[Lafarge and Alliez, 2013]{lafarge_eg13}
Lafarge, F. and Alliez, P. (2013).
\newblock Surface reconstruction through point set structuring.
\newblock {\em Computer Graphics Forum}, 32(2pt2):225--234.

\bibitem[Li et~al., 2016]{li2016boxfitting}
Li, M., Wonka, P., and Nan, L. (2016).
\newblock Manhattan-world urban reconstruction from point clouds.
\newblock In {\em ECCV}, pages 54--69.

\bibitem[Li et~al., 2011]{LWCSCOM11}
Li, Y., Wu, X., Chrysathou, Y., Sharf, A., Cohen-Or, D., and Mitra, N.~J.
  (2011).
\newblock Globfit: Consistently fitting primitives by discovering global
  relations.
\newblock {\em ACM transactions on graphics (TOG)}, 30(4):1--12.

\bibitem[Monszpart et~al., 2015]{monszpart2015rapter}
Monszpart, A., Mellado, N., Brostow, G.~J., and Mitra, N.~J. (2015).
\newblock {RAP}ter: Rebuilding man-made scenes with regular arrangements of
  planes.
\newblock {\em ACM Trans. Graph.}, 34(4):1--12.

\bibitem[Musialski et~al., 2013]{Musialski_cgf13}
Musialski, P., Wonka, P., Aliaga, D.~G., Wimmer, M., Gool, L., and Purgathofer,
  W. (2013).
\newblock A survey of urban reconstruction.
\newblock {\em Comput. Graph. Forum}, 32(6):146–177.

\bibitem[Nan and Wonka, 2017]{nan2017polyfit}
Nan, L. and Wonka, P. (2017).
\newblock Polyfit: Polygonal surface reconstruction from point clouds.
\newblock In {\em 2017 IEEE International Conference on Computer Vision
  (ICCV)}, pages 2372--2380.

\bibitem[Oesau et~al., 2016]{oesau_cgf16}
Oesau, S., Lafarge, F., and Alliez, P. (2016).
\newblock {Planar Shape Detection and Regularization in Tandem}.
\newblock {\em Computer Graphics Forum}, 35(1):203--215.

\bibitem[Schnabel et~al., 2007]{schnabel2007efficient}
Schnabel, R., Wahl, R., and Klein, R. (2007).
\newblock Efficient ransac for point-cloud shape detection.
\newblock {\em Computer graphics forum}, 26(2):214--226.

\bibitem[Shapira et~al., 2008]{shapira_sdf08}
Shapira, L., Shamir, A., and Cohen-Or, D. (2008).
\newblock Consistent mesh partitioning and skeletonisation using the shape
  diameter function.
\newblock {\em The Visual Computer}, 24(4):249–259.

\bibitem[Tulsiani et~al., 2017]{tulsiani2017learning}
Tulsiani, S., Su, H., Guibas, L.~J., Efros, A.~A., and Malik, J. (2017).
\newblock Learning shape abstractions by assembling volumetric primitives.
\newblock In {\em Proceedings of the IEEE Conference on Computer Vision and
  Pattern Recognition}, pages 2635--2643.

\bibitem[V{\'a}rady et~al., 1998]{VBK98}
V{\'a}rady, T., Benko, P., and Kos, G. (1998).
\newblock {Reverse engineering regular objects: simple segmentation and surface
  fitting procedures}.
\newblock {\em Int. J. of Shape Modeling}, 3(4):127--141.

\bibitem[von Luxburg, 2007]{Luxburg07}
von Luxburg, U. (2007).
\newblock A tutorial on spectral clustering.
\newblock {\em Statistics and Computing}, 17:395--416.

\bibitem[Xiao and Furukawa, 2014]{xiao2014}
Xiao, J. and Furukawa, Y. (2014).
\newblock Reconstructing the world's museums.
\newblock {\em International Journal of Computer Vision}, 110(3):243--258.

\end{thebibliography}

\end{document}